\documentclass{article}

\PassOptionsToPackage{numbers, compress}{natbib}

  \usepackage[eandd, final]{neurips_2026}

\usepackage[utf8]{inputenc} 
\usepackage[T1]{fontenc}    
\usepackage{hyperref}       
\usepackage{url}            
\usepackage{booktabs}       
\usepackage{amsfonts}       
\usepackage{nicefrac}       
\usepackage{microtype}      
\usepackage{xcolor}         
\usepackage{graphicx}
\usepackage{amsmath}
\usepackage{subcaption}
\usepackage[labelfont=bf]{caption}
\usepackage{multirow}
\usepackage{enumitem}
\usepackage{float}
\usepackage{tabularx}
\title{PhysMetrics.Weather: An Evaluation Framework for Physical Consistency in ML Weather Models}
\author{%
 Emma Kasteleyn,$^{1,2}$ Timo Maier,$^{2,3}$ Axel Lauer,$^2$ Veronika Eyring,$^{2,4}$ Pierre Gentine$^5$, Ana Lucic$^1$\\
   $^1$ University of Amsterdam, Amsterdam, The Netherlands, 
  $^2$ Deutsches Zentrum f\"{u}r Luft- und  \\ Raumfahrt (DLR), Institut f\"{u}r Physik der Atmosph\"{a}re, Oberpfaffenhofen, Germany, \\
  $^3$ Technical University of Munich, Munich, Germany, \\
  $^4$ University of Bremen, Institute of Environmental Physics (IUP), Bremen, Germany, \\
  $^5$ Department of Earth and Environmental Engineering, Columbia University, New York, NY, USA, 
}

\begin{document}

\maketitle

\begin{abstract}
    Machine learning weather prediction (MLWP) models have achieved impressive forecasting performance at a small fraction of the computational costs required for traditional physics-based methods. 
    However, they are primarily (1) data-driven and (2) evaluated using pixel-wide error metrics (e.g., RMSE), so there are no guarantees that their forecasts are consistent with known physical laws. 
    We introduce \textbf{PhysMetrics.Weather}, an evaluation framework that assesses the physical realism of MLWP models across three types of metrics: conservation, spectral, and dynamical. By quantifying physical realism, this tool guides the development of physics-informed architectures and helps evaluate whether MLWP models are reliable for operational use. Our framework is available on \href{https://github.com/Emmakast/PhysMetrics.Weather}{GitHub}.
\end{abstract}

\section{Introduction}
Modern weather forecasting relies heavily on established numerical weather prediction (NWP) systems like the Integrated Forecasting System (IFS) \citep{ecmwf_ifs_dynamics_2024} and the Global Forecast System (GFS) \citep{noaa2004gfs}. 
They work by iteratively solving complex fluid dynamics equations on massive grids at short time steps, making them computationally expensive.  
These high costs, together with the availability of petabytes of high-quality simulation data, have driven a shift toward machine learning weather prediction (MLWP), which generates forecasts orders of magnitude faster \citep{schultz2021can, pathak_fourcastnet_2022, bi_pangu-weather_2022, lam_graphcast_2023,bodnar_foundation_2025}.

To achieve this speed, MLWP approaches learn weather-relevant atmospheric processes directly from reanalysis datasets such as ERA5 \citep{Hersbach2023ERA5}, which combine observations with traditional numerical simulations. Vision transformers like FourCastNet \citep{pathak_fourcastnet_2022} and Pangu-Weather \citep{bi_pangu-weather_2022}, use the parallelizable nature of self-attention to efficiently process massive spatial grids. Alternatively, graph neural networks (GNNs) such as GraphCast \citep{lam_graphcast_2023} map the atmosphere onto a spherical mesh, allowing for global message passing while avoiding the polar distortions of standard latitude-longitude grids. Extending these architectures, foundation models like Aurora \citep{bodnar_foundation_2024} pre-train on large, heterogeneous datasets to learn general-purpose representations that can be fine-tuned for specific forecasting tasks.

MLWP models predict the next atmospheric state at $t_0 + \Delta t$ from the current state at $t_0$. To generate forecasts, they auto-regressively feed their output back in as the next input. While accurate at short lead times, this iteration exposes a fundamental flaw: deterministic MLWP models progressively smooth the forecast. Standard point-wise loss functions (e.g., MSE, MAE) penalize a spatially shifted feature, like a misplaced storm, twice: once for the miss (a false negative) and again for predicting the storm where none exists (a false positive). To avoid this "double penalty" \citep{hoffman_distortion_1995}, MLWP models tend to predict a smoothed average of possible future states, which often erases fine-scale details and underestimates extreme weather \citep{lam_graphcast_2023, subich_fixing_2025}. Specifically, they struggle to maintain dynamical balance, such as the required coupling between pressure and wind fields to capture cyclone intensities \citep{NorthernHemisphereMidlatitudeCycloneIntensityBiasesinMachineLearningWeatherPredictionModels}. This smoothing also flattens fine-scale convective dynamics, erasing sharp frontal boundaries and vertical wind velocities \citep{pappenberger2024machine}. 
By violating conservation laws, MLWPs can yield solutions that look plausible but are physically inconsistent \citep{bonavita_limitations_2024}.

Current MLWP benchmarks (e.g., WeatherBench \citep{rasp_weatherbench_2020,rasp_weatherbench_2024,weatherbench_x_2025}) provide standardized datasets and metrics for comparing MLWP models against operational systems like the IFS \cite{ecmwf_ifs_dynamics_2024}. 
However, they focus on evaluating predictive skill via metrics like latitude-weighted RMSE, which rewards spatial smoothing. While these benchmarks are excellent for assessing point-wise predictive accuracy, they offer limited insight into whether a model maintains physical consistency when generating forecasts.

To address this gap, we introduce \textbf{PhysMetrics.Weather}, a framework designed to complement traditional error metrics by evaluating the physical consistency of MLWP models. While MLWP models offer computational advantages, assessing their reliability for forecasting and climate science requires verifying that they respect physical laws over the entire forecast period \citep{reichstein2019deep, PhysRevLett.126.098302}. PhysMetrics.Weather evaluates models across three physical metric types: global conservation budgets (mass and energy), spectral energy distribution (preservation of spatial variance), and multi-variable dynamical balances. PhysMetrics.Weather provides machine learning scientists with standardized insights into whether their models are physically realistic.

\section{Related Work}
\begin{table}[h]
\centering
\caption{Comparison of MLWP benchmarks. \textbf{PhysMetrics.Weather} is the first to provide a unified evaluation of conservation budgets, spectral energy distribution, and dynamical balance (e.g., geostrophic and hydrostatic equilibria). Acronyms: ACC (Anomaly Correlation Coefficient), CRPS (Continuous Ranked Probability Score).}
\label{tab:benchmark_comparison}
\resizebox{\textwidth}{!}{%
\begin{tabular}{@{}l c c c c c c @{}}
\toprule
\textbf{Benchmark} & \textbf{Primary Focus} & \textbf{Statistical} & \textbf{Probabilistic} & \textbf{Conservation} & \textbf{Structural} & \textbf{Dynamical} \\ 
 & & (RMSE/ACC) & (CRPS/Spread) & (Budgets) & (Spectral) & (Balances) \\ \midrule
WeatherBench \citep{rasp_weatherbench_2020} & Predictive Skill & \checkmark & $\times$ & $\times$ & $\times$ & $\times$ \\
WeatherBench 2 \citep{rasp_weatherbench_2024} & Operational Standard & \checkmark & \checkmark & $\times$ & $\times$ & $\times$ \\
WeatherBench-X \citep{weatherbench_x_2025} & Modular Evaluation & \checkmark & \checkmark & $\times$ & $\times$ & $\times$ \\
ExtremeWB \citep{noauthor_brightbandtechextremeweatherbench_2026} & Rare Events & \checkmark & $\times$ & $\times$ & $\times$ & $\times$ \\
ClimateLearn \citep{nguyen2023climatelearnbenchmarkingmachinelearning} & Data Standardization & \checkmark & $\times$ & $\times$ & $\times$ & $\times$ \\
ChaosBench \citep{nathaniel_chaosbench_2024} & Dynamics/Chaos & \checkmark & $\times$ & $\times$ & \checkmark & $\times$ \\ \midrule
\textbf{PhysMetrics.Weather (Ours)} & \textbf{Phys. Consistency} & $\times$ & $\times$ & \checkmark & \checkmark & \checkmark \\
\bottomrule
\end{tabular}%
}
\end{table}
Table \ref{tab:benchmark_comparison} summarizes commonly used MLWP benchmarks. WeatherBench \citep{rasp_weatherbench_2020} and its successors, WeatherBench 2 \citep{rasp_weatherbench_2024} and WeatherBench-X \citep{weatherbench_x_2025}, established a widely accepted baseline for evaluating MLWP models. They provide a common reference point for the research community by standardizing the use of metrics like latitude-weighted RMSE and spread-skill ratios \citep{gneiting2007strictly, fortin2014why}. While these frameworks are the standard for quantifying point-wise errors and probabilistic spread, they do not diagnose whether a model's predicted atmospheric state remains physically consistent over time.

Recent benchmarks address specific forecasting challenges. ExtremeWeatherBench \citep{noauthor_brightbandtechextremeweatherbench_2026} evaluates rare, high-impact events, while ClimateLearn \citep{nguyen2023climatelearnbenchmarkingmachinelearning} offers a reproducible framework by standardizing data pipelines and evaluation protocols across diverse datasets (e.g., ERA5 \citep{Hersbach2023ERA5}, CMIP6 \citep{cmip6_highresmip_2018}). ChaosBench \citep{nathaniel_chaosbench_2024}, our closest precursor, shifts the focus towards atmospheric dynamics, using spectral diagnostics to test whether models reproduce a realistic energy distribution of the atmosphere. 
While ChaosBench takes an important step towards evaluation of physical consistency, its scope is limited to spectral properties (the distribution of energy across spatial scales).

Beyond standardized benchmarks, the field has increasingly focused on enforcing physical consistency directly within the modeling pipeline, acknowledging that models can yield predictions that are statistically plausible but physically inconsistent \citep{bonavita_limitations_2024}. Soft constraint approaches add loss penalties to preserve fine-scale detail \citep{dunstan_fastnet_2025, saccardi_assessing_2025, subich_fixing_2025}, suppress global budget drifts \citep{sha_improving_2025-1}, or enforce multivariate consistency through latent-space constraints \citep{fan2025incorporating}. Conversely, hard constraint approaches embed differentiable physics solvers directly into the architecture e.g., hybrid models like NeuralGCM \citep{kochkov_neural_2024} couple traditional, differentiable physics solvers with neural network sub-modules, combining NWP stability with ML computational efficiency. Alternatively, models like ClimODE \citep{verma_climode_2024} use neural ordinary differential equations to represent atmospheric evolution as a continuous physical process.

To evaluate physically constrained MLWP designs, researchers typically rely on custom diagnostics tailored to their specific architectural objectives. For instance, to show their models overcome the smoothing effect, authors use spectral analysis to verify energy retention at fine scales \citep{subich_fixing_2025, saccardi_assessing_2025, dunstan_fastnet_2025}. To detect localized grid-cell noise, some researchers evaluate spatial gradients to ensure wind fields maintain geostrophic balance with the atmospheric pressure gradient \citep{saccardi_assessing_2025, dunstan_fastnet_2025}. Similarly, other studies validate embedded hydrostatic constraints by measuring layer-wise thermodynamic imbalance \citep{subramaniam_imposing_2025}, or calculate global budgets to show that loss penalties prevent long-term mass and energy drift \citep{sha_improving_2025-1}.

However, since researchers use disparate metrics to validate their specific models, it is challenging to objectively compare physical consistency across different models. By unifying conservation, spectral, and dynamical metrics into a single framework, PhysMetrics.Weather provides a common evaluation framework to assess whether MLWP models preserve physical laws over the entire forecast period. PhysMetrics.Weather is available for download as an open-source repository on \href{https://github.com/Emmakast/PhysMetrics.Weather}{GitHub}.

\section{Methodology}\label{sec:methodology}
\begin{figure}[t]
    \centering
\includegraphics[width=\linewidth]{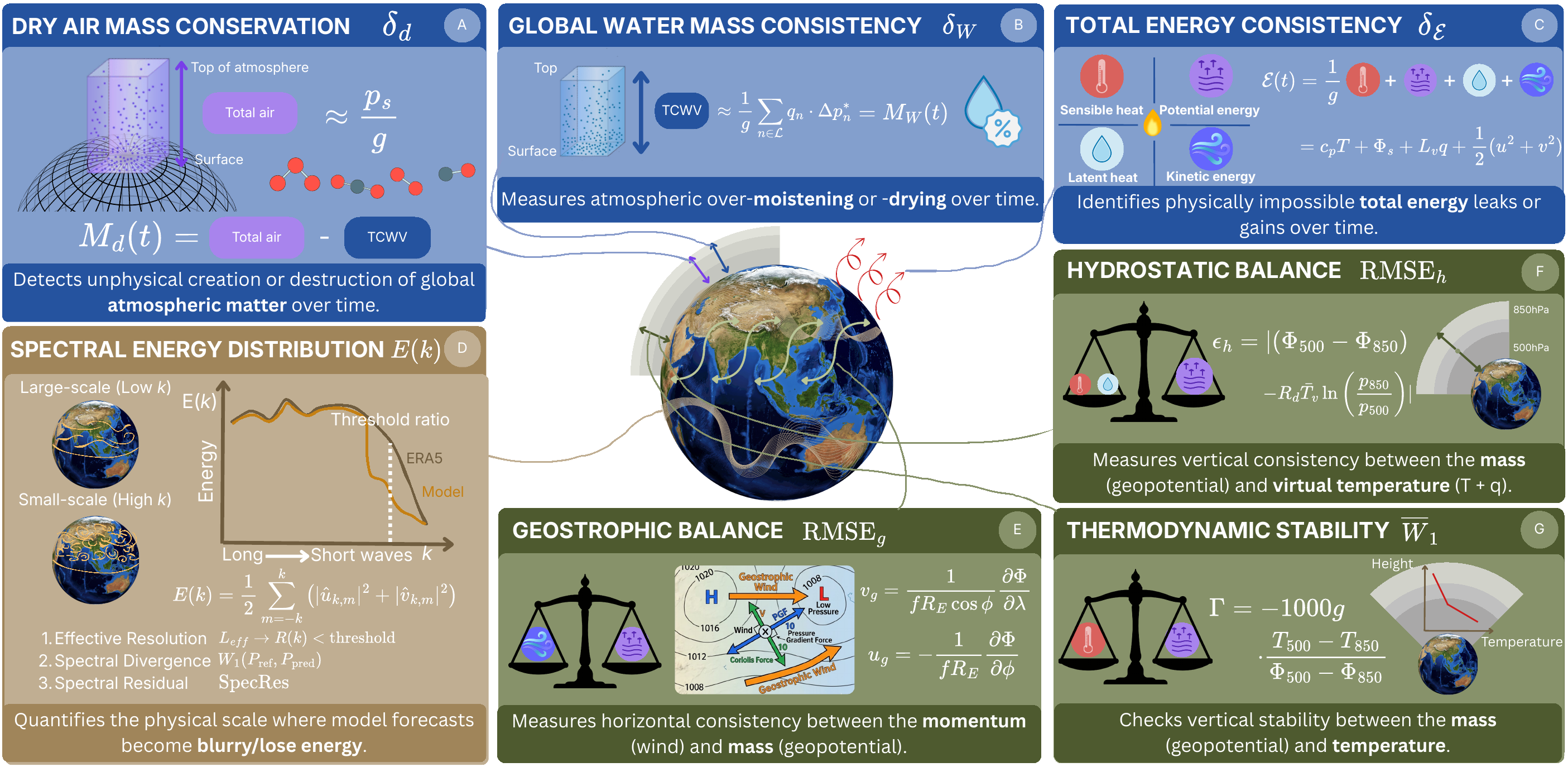}
    \caption{\textbf{Overview of PhysMetrics.Weather.} PhysMetrics.Weather evaluates models using nine metrics across three metric types: conservation of mass and energy (blue), spectral energy distribution (beige), and adherence to dynamical balance (green). Metric definitions are detailed in Table~\ref{tab:metrics_summary}. \textbf{A. Dry Air Mass Conservation:} tracks global dry air mass. 
    \textbf{B. Global Water Mass Consistency:} tracks the atmospheric water budget. 
    \textbf{C. Total Energy Consistency:} estimates the global energy budget across sensible, potential, latent, and kinetic states. 
    \textbf{D. Spectral Energy Distribution:} evaluates the spectral distribution of wind energy across spatial frequencies (using three metrics). 
    \textbf{E. Geostrophic RMSE:} evaluates if models reproduce the relationship between wind speeds and pressure gradients (the Coriolis effect). 
    \textbf{F. Hydrostatic RMSE:} evaluates atmospheric stability by checking the temperature of the air between two pressure levels.
    \textbf{G. Thermodynamic Stability:} evaluates the plausibility of vertical temperature gradients (lapse rates) across climate zones.}
    \label{fig:graphic}
\end{figure}

Here, we detail the methodology behind \textbf{PhysMetrics.Weather} (Figure \ref{fig:graphic}). Extending beyond standard error evaluations (Appendix \ref{app:RMSE}), our framework captures an independent axis of model performance by evaluating physical consistency metrics across three metric types: (1) conservation of mass and energy (Section \ref{sec:conservation}), (2) spectral energy distribution (Section \ref{sec:structure}), and (3) dynamical balances (Section \ref{sec:balance}). While their relative importance varies by downstream task, monitoring all three types is important for model reliability.

Our evaluation uses the publicly available prediction datasets from WeatherBench 2 \citep{rasp_weatherbench_2024}, focusing on state-of-the-art MLWP and NWP models with evaluations available for the 2020 test set: Pangu-Weather \citep{bi_pangu-weather_2022}, GraphCast \citep{lam_graphcast_2023}, FuXi \citep{sun_fuxi_2024}, NeuralGCM \citep{kochkov_neural_2024}, and IFS HRES \citep{ecmwf_hres_2024}. For a fair comparison, we evaluate the same subset of atmospheric variables available from all models: specific humidity ($q$), geopotential ($\Phi$), air temperature ($T$), horizontal wind components ($u, v$), and surface pressure ($p_s$).

We evaluate models using one daily initial condition (00:00 UTC) throughout the year 2020 (366 days). For each model and initial day, we track the 10-day forecast trajectory at the model's native temporal resolution (typically 6 hours) and native vertical resolution (either 13 \citep{bi_pangu-weather_2022, sun_fuxi_2024} or 37 \citep{lam_graphcast_2023, kochkov_neural_2024, ecmwf_hres_2024} pressure levels).

Our framework aggregates predictions for all grid cells into global budgets using a weighted sum, denoted as $\langle \cdot \rangle_\mathcal{D}$, over the model's spherical grid $\mathcal{D}$. This operator computes an area-weighted spatial sum, $\langle \cdot \rangle_\mathcal{D}=\sum_{i,j\in\mathcal{D}}A_{i,j}X_{i,j}$, where each cell $(i,j)$ contributes proportionally to its true physical surface area $A_{i,j}$, preventing polar distortion (Appendix \ref{app:area_weights}).

For 3D atmospheric quantities, we perform vertical column integration across the given pressure levels, using surface pressure ($p_s$) as the lower boundary. Depending on the model, $p_s$ is provided directly, estimated from mean sea level pressure, or otherwise taken from the reference data. We integrate across the vertical layers using a trapezoidal rule. The integration weights ($\Delta p_n^*$) act as a mask assigning zero weight to pressure levels below the Earth's surface (Appendix \ref{app:integration_global}).

A reference dataset serves as a proxy for the true atmospheric state. While traditionally used to measure predictive accuracy, we use references like ERA5 \citep{Hersbach2023ERA5} and IFS HRES analysis \citep{ecmwf_hres_2024} to evaluate physical consistency. Although PhysMetrics.Weather supports any atmospheric reference dataset, we use the ERA5 reanalysis \citep{Hersbach2023ERA5} in this work. As the main training corpus for most MLWP models \citep{bi_pangu-weather_2022, lam_graphcast_2023, sun_fuxi_2024, kochkov_neural_2024}, evaluating against ERA5 explicitly tests if models preserve the physics implicit in their training data. To ensure our findings are not solely dependent on reanalysis data, we also validate our results using the IFS HRES analysis (Appendix \ref{app:ifs}), which is a reference that operational forecasters trust and rely on its degree of physical consistency.

\subsection{Metrics for Evaluating Conservation of Mass and Energy}
\label{sec:conservation}
We adopt the global mass calculation from \citet{sha_improving_2025-1}, and extend it by introducing a drift metric to quantify physical consistency. Using this approach, we test whether MLWP models implicitly learn and maintain the conservation of mass and energy by calculating the following three metrics:

\textbf{Dry Air Mass Conservation:} 
Over weather prediction timescales, dry air mass is conserved \citep{trenberth2005mass}. While reanalysis reference datasets may show minor mass fluctuations due to data assimilation, trends in predicted total air mass indicate unrealistic sources or sinks of matter. Assuming hydrostatic balance, we obtain the global dry air mass $M_d(t)$ by calculating the surface integral over the entire globe: we divide the surface pressure ($p_s$) by gravitational acceleration ($g$) to derive total atmospheric mass, then subtract the total column water vapor (TCWV, the vertically integrated specific humidity $q$) to convert to dry air.

We report the \textit{relative drift rate} ($\delta_d$), the percentage of mass change per day. We use the the slope of the least-squares regression ($m_d$) fit to the trajectory $M_d^{\text{pred}}(t)$ over the prediction time period $\mathcal{T} = [t_{0}, t_{\text{eval}}]$, and normalize the result by the initial value. A realistic change rate should be close to zero, indicating no significant sources or sinks of dry air.
$$M_d(t) = \left\langle \frac{p_s}{g} - \text{TCWV} \right\rangle_\mathcal{D},\ \ \delta_d(\mathcal{T}) = \frac{m_d(\mathcal{T})}{M_d^{\text{pred}}(t_{0})} \times 100$$

\textbf{Global Water Mass Consistency:} Unlike dry air, atmospheric water mass fluctuates via evaporation and precipitation. A physical model should balance these sources and sinks over longer time periods, avoiding systematic moistening or drying of the atmosphere. We define global water mass $M_w(t)$ as TCWV integrated over the whole surface area of the globe.

Since reanalysis data does not have a perfectly closed hydrological or thermodynamic budget (due to assimilation), we compute the \textit{anomaly drift rate} ($\delta_w$) to measure how much the predicted mass trend ($m_w^{\text{pred}}$) deviates from the reference trend ($m_w^{\text{ref}}$) over the same time period $\mathcal{T} = [t_{0}, t_{\text{eval}}]$. A model's anomaly drift rate should be close to zero, meaning trends in the total atmospheric water mass caused by evaporation and precipitation over the prediction time period match the trends of the reference dataset.

$$M_w(t) = \langle \text{TCWV}_\text{pred} \rangle_\mathcal{D}, \ \ \delta_w(\mathcal{T}) = \left( \frac{m_w^{\text{pred}}(\mathcal{T})}{M_w^{\text{pred}}(t_{0})} - \frac{m_w^{\text{ref}}(\mathcal{T})}{M_w^{\text{ref}}(t_{0})} \right) \times 100$$

\textbf{Total Energy Consistency:} The global energy budget consists of the thermal energy (sensible heat, $c_p T$), gravitational energy (potential energy, $\Phi_s$), moisture-related energy (latent heat, $L_v q$), and momentum (kinetic wind energy (KE), $\frac{1}{2}(u^2+v^2)$). While energy constantly shifts between these forms, the global energy budget should remain balanced relative to natural fluxes; for example, wind acceleration without a corresponding temperature or pressure change implies violations of the energy conservation. We estimate the total global energy $\mathcal{E}(t)$ by integrating the energy sum across all grid cells and all pressure levels. The constants ($c_p, L_v, g$) are defined in Appendix \ref{app:constants}. 

Applying the same approach as for the total water content, we compute the \textit{anomaly drift rate} ($\delta_\mathcal{E}$) to measure deviation from the reference data energy trend. A physically consistent model needs to preserve total energy, yielding an anomaly drift of close to zero.

$$\mathcal{E}(t) = \frac{1}{g} \left\langle \int_0^{p_s} \left(c_p T + \Phi_s + L_v q + \frac{1}{2}(u^2+v^2)\right) dp \right\rangle_\mathcal{D}, \ \ \delta_\mathcal{E}(\mathcal{T}) = \left( \frac{m_\mathcal{E}^{\text{pred}}(\mathcal{T})}{\mathcal{E}^{\text{pred}}(t_{0})} - \frac{m_\mathcal{E}^{\text{ref}}(\mathcal{T})}{\mathcal{E}^{\text{ref}}(t_{0})} \right) \times 100$$

\subsection{Metrics for Evaluating Spectral Energy Distribution}\label{sec:structure}
To quantify the smoothing effect, we apply a spherical harmonic transform (SHT) \citep{subich_fixing_2025,kochkov_neural_2024} (the spherical equivalent of a 2D Fourier transform) to decompose the predicted horizontal wind fields into spatial wavenumbers ($k$). Low and high wavenumbers correspond to large-scale (e.g., Rossby waves) and fine-scale (e.g., frontal boundaries) phenomena, respectively. We evaluate the kinetic energy (KE, $E$) spectrum of the wind field at 500 hPa. At this altitude in the mid-troposphere, there are no effects from friction at the Earth's surface. The winds at 500 hPa are characteristic for the large-scale flow that determines the movement of extratropical cyclones, and thus, surface weather patterns (see Appendix \ref{app:spectral_ablation} for 850 hPa KE and humidity analyses). The KE spectrum depends on the zonal (eastward, $u$) and meridional (northward, $v$) wind components. By summing the complex SHT coefficients ($\hat{u}_{k,m}$ and $\hat{v}_{k,m}$) over all zonal wavenumbers ($m$) for a given total wavenumber ($k$), the KE at spatial scale $k$ is calculated as:

$$E(k) = \frac{1}{2} \sum_{m=-k}^{k} \left( |\hat{u}_{k,m}|^2 + |\hat{v}_{k,m}|^2 \right)$$ 

\textbf{Effective Resolution ($L_{eff}$):} This evaluates a model's spatial resolution by measuring the retained fraction of reference KE at different spatial scales. Since deterministic MLWP models tend to smooth fine details over time, we identify the exact physical wavelength where this smoothing effect starts to decay the model's structural variance. We calculate the spectral energy retention ratio $R(k)$ at each wavenumber for a given lead time by comparing the predicted energy spectrum to the reference spectrum, with both spectra averaged across all evaluation dates:

$$R(k) = \frac{\bar{E}_\text{pred}(k)}{\bar{E}_\text{ref}(k)}$$

The \textit{effective resolution} ($L_{eff}$) is the wavelength where the retention ratio drops below 50\% ($R(k) < 0.5$) for five consecutive wavenumbers. Ideally, $L_{eff}$ matches the model's native resolution, indicating good energy retention at fine scales (see Appendix \ref{app:thresholding} for alternative thresholds).

\textbf{Spectral Residual:} This measures the absolute error magnitude across the energy spectrum. Atmospheric energy follows a power-law cascade, meaning large-scale waves contain orders of magnitude more energy than fine-scale features \citep{nastrom1985climatology}. A simple RMSE would underrepresent errors at fine scales. To penalize high-frequency loss equally across this cascade up to the maximum spatial wavenumber $K$, we adapt the logarithmic \textit{spectral residual} introduced by \citet{nathaniel_chaosbench_2024}, computing it as the RMSE of the log-transformed spectrum. A physically realistic model should have an RMSE close to zero, demonstrating that its energy distribution accurately matches the reference data across all spatial scales.

$$\text{SpecRes} = \sqrt{ \frac{1}{K+1} \sum_{k=0}^{K} \left( \log E_\text{pred}(k) - \log E_\text{ref}(k) \right)^2 }$$

\textbf{Spectral Divergence:} This assesses the shape of the energy cascade rather than its absolute magnitude. Disproportionate energy in low frequencies indicates smoothing, while high-frequency spikes at fine spatial scales suggest gridpoint artifacts \citep{subich_fixing_2025}.  
We treat the energy spectrum as a probability distribution by normalizing the energy at each wavenumber by the total sum of the spectrum.

Inspired by \citet{nathaniel_chaosbench_2024}, we compute the \textit{spectral divergence} \citep{nathaniel_chaosbench_2024} using the 1-Wasserstein distance ($W_{1}$) between the reference and predicted distributions. Ideally, the model should achieve a spectral divergence near zero, showing that it allocates energy proportionally across all spatial scales without excessive smoothing.

$$P(k) = \frac{E(k)}{\sum E(j)}, \ \ \text{SpecDiv} = W_{1}(P_\text{ref}, P_\text{pred}) = \sum_{k=0}^{K} \left| \sum_{j=0}^{k} P_\text{ref}(j) - \sum_{j=0}^{k} P_\text{pred}(j) \right|$$

\subsection{Metrics for Evaluating Dynamical Balance}\label{sec:balance}
We also evaluate whether models maintain physical consistency between distinct output variables (e.g., coupling wind, temperature, and pressure fields). Because the real atmosphere naturally deviates from idealized equilibria (e.g., in storms) and exhibits diverse vertical temperature gradients \citep{holton2012dynamic}, a realistic model must reproduce the natural variability of the reference dataset. We quantify this in two ways: for physical equilibria (geostrophic and hydrostatic balances), we compute the excess imbalance ($\Delta \text{RMSE}$) between the model and reference residuals; for vertical thermodynamic stability, we evaluate the preservation of atmospheric lapse rate distributions.

\textbf{Geostrophic Balance:} Large-scale mid-latitude flow is governed by geostrophic balance, where Coriolis and pressure gradient forces align \citep{holton2012dynamic}. A physically realistic model must implicitly couple mass and momentum fields; for example, predicting a deep low-pressure system without generating corresponding cyclonic winds violates physical consistency \citep{bonavita_limitations_2024}. We compute the geostrophic wind components $(u_g, v_g)$ derived from the spatial gradient of the predicted geopotential $\Phi$, where $\phi$ and $\lambda$ are latitude and longitude, $R_E$ is Earth's radius, and $f$ is the Coriolis parameter.

We evaluate how well the model couples its mass and momentum fields by computing the vector RMSE between the predicted wind ($u, v$) and the theoretical geostrophic wind. We exclude the equatorial region ($|\phi| < 10^\circ$), where geostrophic balance is invalid due to a weak Coriolis force, as well as the poles ($|\phi| \geq 89.9^\circ$) to avoid numerical singularities. Given that there are usually (small) deviations from this equilibrium (ageostrophic flow), a realistic model should not force this absolute residual to zero \citep{holton2012dynamic}. Instead, it should yield an \textit{excess geostrophic imbalance} ($\Delta \text{RMSE}_g$) near zero, i.e. maintaining the natural ageostrophic characteristics of the reference dataset.

$$u_g = -\frac{1}{f R_E} \frac{\partial \Phi}{\partial \phi}, \quad v_g = \frac{1}{f R_E \cos\phi} \frac{\partial \Phi}{\partial \lambda}, \ \ \Delta \text{RMSE}_g = \text{RMSE}_g^\text{pred} - \text{RMSE}_g^\text{ref}$$

\textbf{Hydrostatic Balance:} To a large degree, the upward pressure gradient force in the atmosphere is balanced by the downward-directed gravitational acceleration. Hydrostatic balance is generally an accurate approximation for grid-scale air columns typical of global weather forecast and reanalysis models \citep{holton2012dynamic}. As the vertical spatial distance (geopotential height difference) between two vertical levels is determined by the temperature and moisture of the air in-between, a model predicting e.g. a temperature spike without proportionally increasing the geopotential distance would violate this hydrostatic balance. We measure this constraint using the residual ($\epsilon_h$) of the hypsometric equation, which relates an atmospheric pressure ratio to the equivalent thickness of an atmospheric layer using the dry air gas constant ($R_d$) and the layer's mean virtual temperature, $T_v = T(1 + 0.6078q)$.

We assess thermodynamic consistency (hydrostatic balance) by computing the area-weighted RMSE of this hydrostatic residual. To account for natural non-hydrostatic vertical motions (e.g., thunderstorms), we report the \textit{excess hydrostatic imbalance} ($\Delta \text{RMSE}_h$) compared to the reference data. Note that this reference imbalance primarily stems from vertical interpolation and data assimilation artifacts, rather than explicitly resolved non-hydrostatic convection \citep{Hersbach2023ERA5}. A model should achieve an excess imbalance of close to zero, preserving vertical thermodynamic consistency while allowing for natural, non-hydrostatic vertical dynamics.

$$\epsilon_h = \left| (\Phi_{500} - \Phi_{850}) - R_d \bar{T}_v \ln\left(\frac{p_{850}}{p_{500}}\right) \right|, \ \ \Delta \text{RMSE}_h = \text{RMSE}_h^\text{pred} - \text{RMSE}_h^\text{ref}$$

\textbf{Vertical Thermodynamic Stability:} The atmospheric lapse rate ($\Gamma$) is the rate at which temperature decreases with height, which is a measure for the vertical stability \citep{Stone_and_Carlson_1979}. The lapse rate can be used to check inter-variable and inter-layer relations between temperature and geopotential. To evaluate whether models produce realistic atmospheric stability, we calculate the distribution of lapse rates between 500 hPa and 850 hPa for specific regions (in units of K/km):

$$\Gamma = -g \frac{T_{500}-T_{850}}{\Phi_{500}-\Phi_{850}}\times 1000$$

We evaluate lapse rates in the Tropics ($30^\circ\text{S}$–$30^\circ\text{N}$) and both mid-latitudes ($30^\circ$–$60^\circ$ N/S). For each region, we compute the area-weighted 1-Wasserstein distance ($W_1$) between the model's lapse rate distribution and the reference data. The simple average of these regional distances yields the \textit{mean lapse rate Wasserstein} ($\overline{W}_1$; Appendix \ref{app:wasserstein}). An ideal $\overline{W}_1=0$ indicates the MLWP model preserves reference atmospheric variance in lapse rates rather than collapsing predictions to a smoothed mean.

\section{Results}
We report results for Pangu-Weather \citep{bi_pangu-weather_2022}, GraphCast \citep{lam_graphcast_2023}, FuXi \citep{sun_fuxi_2024}, NeuralGCM \citep{kochkov_neural_2024}, and IFS HRES \citep{ecmwf_hres_2024} on our three types of metrics below. 

\subsection{Conservation}
\begin{figure*}[htpb]
    \centering
    
    \begin{subfigure}{\textwidth}
        \centering
        \includegraphics[width=0.93\linewidth]{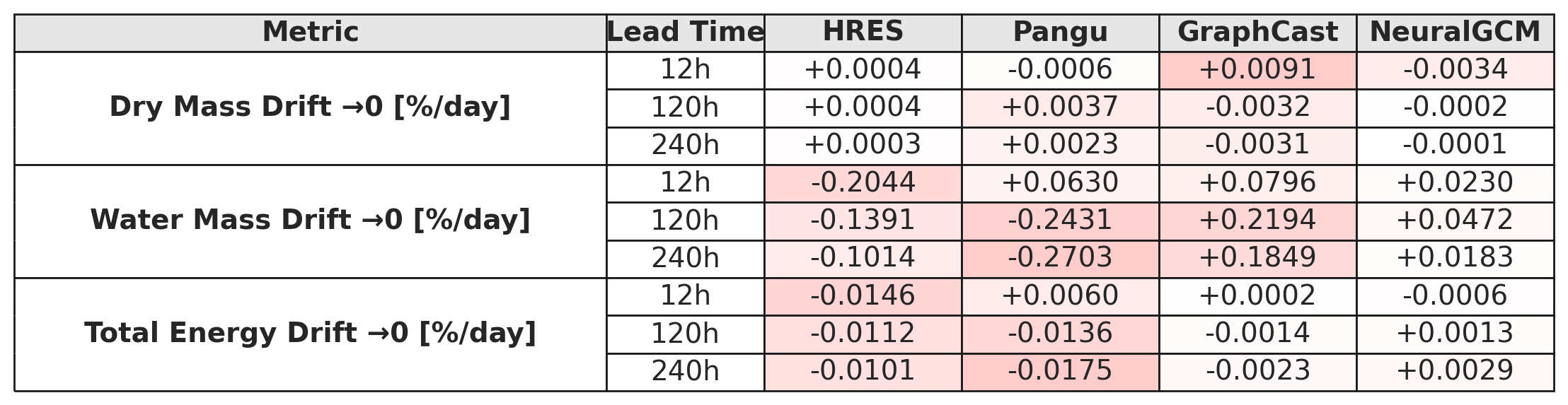} 
    \end{subfigure}
     \begin{subfigure}{\textwidth}
        \centering
        \includegraphics[width=0.93\linewidth]{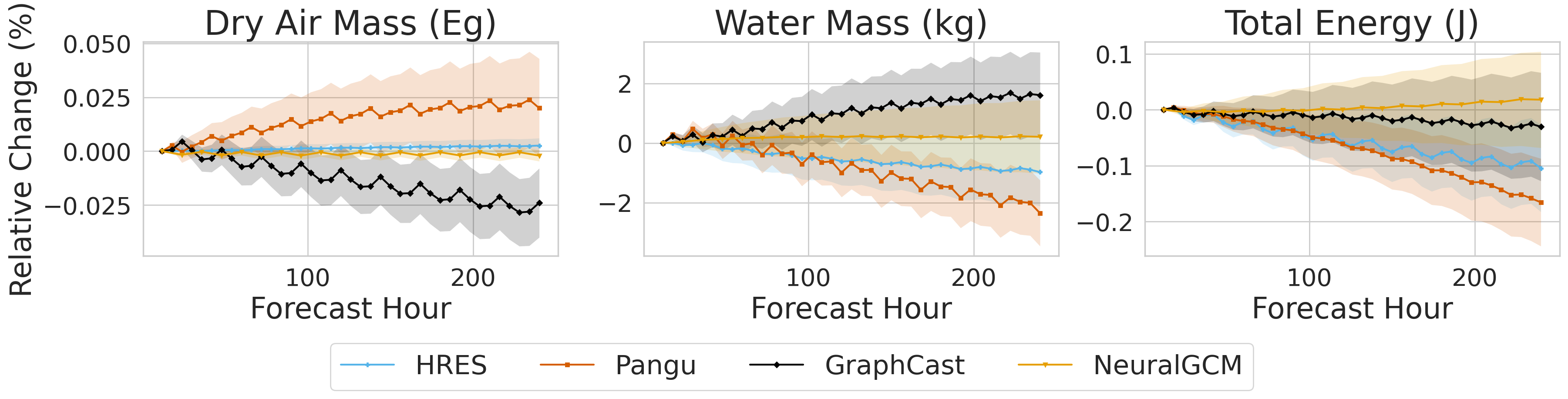}
    \end{subfigure}\hfill
    
    \caption{\textbf{Global mass and energy conservation metrics calculated over a 240-hour prediction time period.} \textbf{Top:} Summary of conservation metrics. \textbf{Bottom Left:} Relative Dry Mass drift. \textbf{Bottom Middle:} Anomaly Water Mass drift. \textbf{Bottom Right:} Anomaly Total Energy drift. Shading indicates the $\pm 1\sigma$ over 2020. Constrained models like NeuralGCM show conservation over extended time periods, while data-driven MLWP models (e.g., GraphCast, Pangu-Weather) show deviations.}
    \label{fig:conservation_metrics}
\end{figure*}

We first examine whether models conserve global mass and energy over the 240-hour prediction period (Figure \ref{fig:conservation_metrics}). FuXi \citep{sun_fuxi_2024} is excluded because specific humidity is unavailable in its WeatherBench 2 outputs. The near-zero \textit{relative} and \textit{anomaly drift} rates demonstrate the stability of NeuralGCM's PDE solver (using ERA5 surface pressure).

The figure also shows that MLWP models struggle to maintain conservation of mass and energy over extended periods. When interpreting these deviations, it is important to note that the ERA5 reference has minor budget imbalances; thus, the models are failing to match the natural variability and assimilation-driven trends of the proxy atmosphere, rather than a state of perfect conservation. GraphCast loses dry air mass while simultaneously generating water mass, whereas Pangu-Weather dries the atmosphere. Finally, the IFS HRES model shows a slight loss of both water mass and energy; this is expected, as operational NWPs are primarily optimized for short- to medium-range predictive accuracy rather than long-term climatological budget conservation. 

\subsection{Energy Spectra}

\begin{figure*}[htpb]
    \centering

    \begin{subfigure}{\textwidth}
        \centering
        \includegraphics[width=0.95\linewidth]{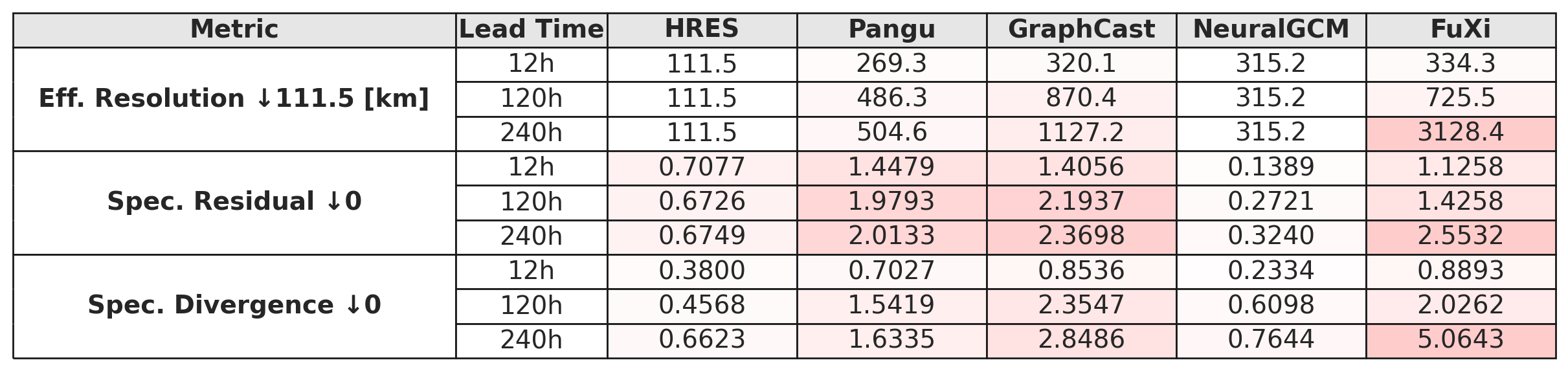} 
    \end{subfigure}

     \begin{subfigure}{\textwidth}
        \centering
        \includegraphics[width=0.95\linewidth]{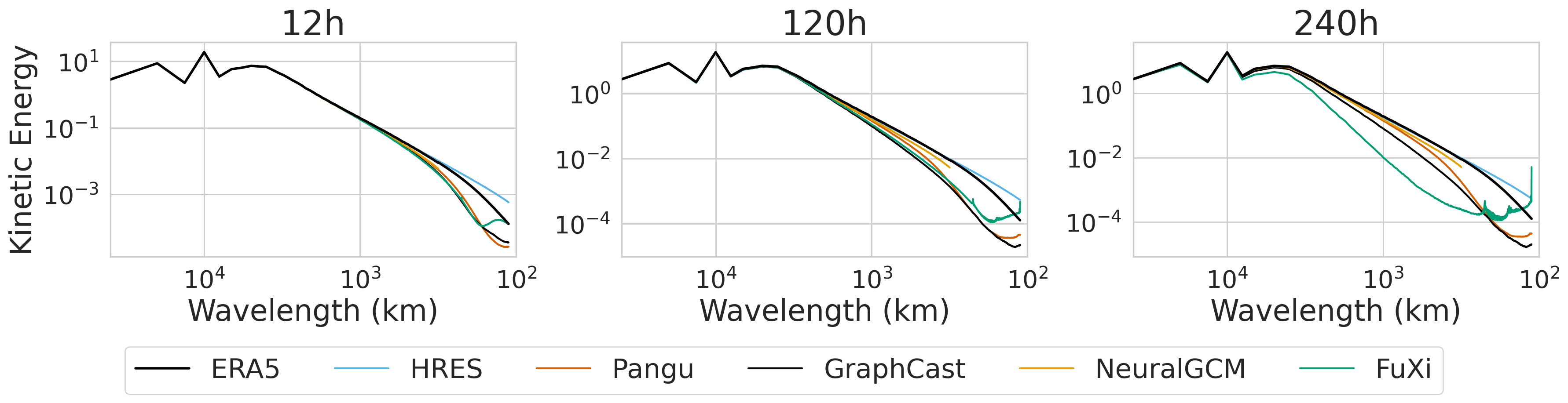}
    \end{subfigure}\hfill
    
    \caption{\textbf{KE spectra at 500 hPa.} \textbf{Top:} Spectral metrics summary. \textbf{Bottom:} Spectra at 12h, 120h, and 240h lead times. Native resolutions: 0.25° (111.5 km), except NeuralGCM (0.7°, 315.2 km). Over extended forecasts, data-driven MLWPs exhibit spatial smoothing or artificial noise, whereas hybrid architectures (e.g., NeuralGCM) preserve realistic energy distributions.}
    \label{fig:combined_spectra}
\end{figure*}

To assess whether models preserve fine-scale atmospheric dynamics, we examine the KE spectra at 500 hPa across lead times (Figure \ref{fig:combined_spectra}). At short lead times (12h), all models align relatively well with the reference (ERA5) baseline. The IFS HRES retains even more energy at fine scales than ERA5, reflecting the difference between ERA5 reanalysis and operational IFS HRES analysis.

At extended lead times (e.g., 120h and 240h), data-driven MLWP models deviate from the reference spectrum. GraphCast and Pangu-Weather suffer energy loss at fine spatial scales over time. This spatial smoothing is quantified by a decrease in their \textit{effective resolution} and an increase in their \textit{spectral residual}. In contrast, FuXi generates excess fine-scale energy, maybe indicating the presence of artificial grid-point noise, a behavior penalized heavily by the \textit{spectral divergence} metric. Both NeuralGCM and IFS HRES maintain stable spectral energy distributions across all lead times.

\subsection{Dynamical Balance}
\begin{figure*}[htpb]
    \centering
    
    \begin{subfigure}{\textwidth}
        \centering
        \includegraphics[width=0.95\linewidth]{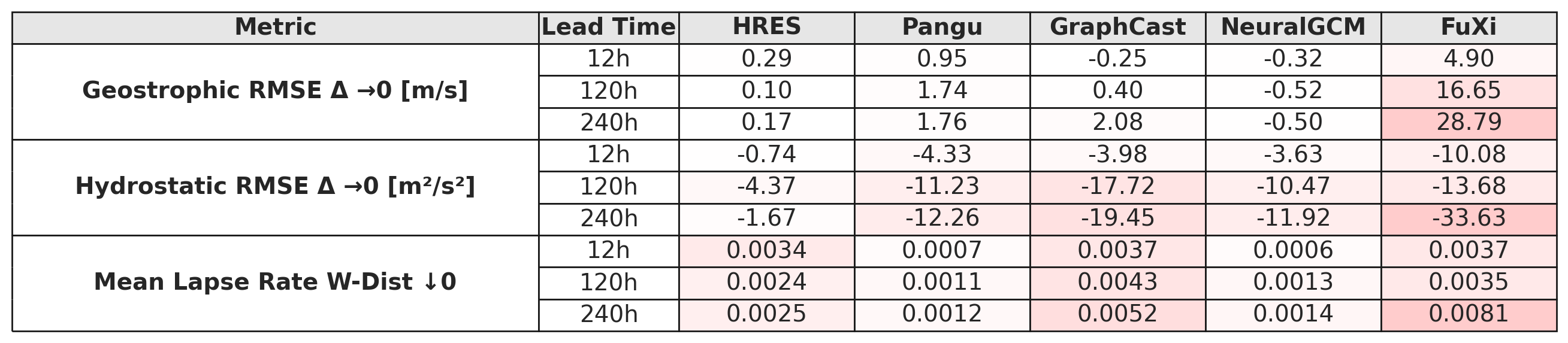} 
    \end{subfigure}

     \begin{subfigure}{\textwidth}
        \centering
        \includegraphics[width=0.95\linewidth]{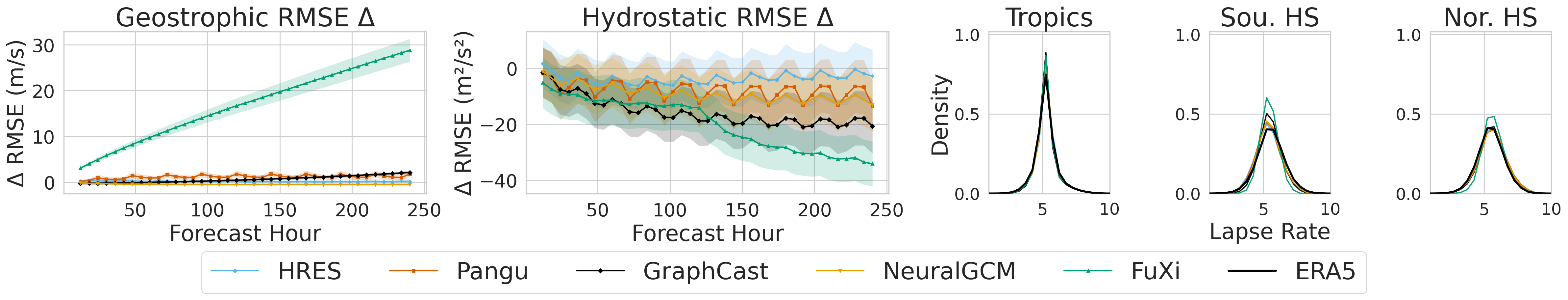}
    \end{subfigure}
    
    \caption{\textbf{Adherence to geostrophic wind balance, hydrostatic equilibrium and realistic atmospheric stability over a 240-hour prediction time period.} \textbf{Top:} Summary of dynamical balance metric. \textbf{Bottom Left:} Excess Geostrophic Imbalance. \textbf{Bottom Middle:} Excess Hydrostatic Imbalance (for FuXi, $T \approx T_v$). \textbf{Bottom Right:} Lapse rate distributions for three regions at 240h lead time. Shading in time series indicates $\pm 1$ standard deviation over 2020.}
    \label{fig:RMSE}
\end{figure*}

Finally, we assess whether models maintain physically consistent relationships between their predicted wind, temperature, and pressure fields (Figure \ref{fig:RMSE}). Because the reference data reflects both natural atmospheric variability (e.g., ageostrophic flow) and inherent structural properties (e.g., vertical interpolation artifacts), a realistic prediction model should match the ERA5 reference residual rather than minimizing the absolute error to zero.

For geostrophic balance, HRES, GraphCast, and Pangu-Weather follow the reference data. However, FuXi's \textit{excess geostrophic imbalance} drifts steadily, showing a decoupling of its momentum and mass fields at extended lead times. NeuralGCM is unique in maintaining a consistently lower geostrophic RMSE than the ERA5 reference itself.

For hydrostatic balance, MLWP models show a decreasing RMSE over time, dropping below the ERA5 reference. This negative \textit{excess hydrostatic imbalance} suggests fine-scale smoothing, resulting in less extreme, artificially balanced states that lack natural, chaotic atmospheric variability.

Low \textit{mean lapse rate Wasserstein} scores indicate that all models broadly reproduce the ERA5 reference. However, MLWP models struggle with the Southern Hemisphere distribution at 240h lead times. This aligns with WeatherBench 2 findings \citep{rasp_weatherbench_2024}, maybe reflecting the sparse observational data in this region that limits training and assimilation constraints, leading to higher forecast uncertainty. Furthermore, while Pangu-Weather and NeuralGCM appear to outperform HRES in thermodynamic stability, Appendix \ref{app:ifs} demonstrates this is an evaluation artifact: evaluating HRES against its own initialization (IFS HRES analysis) rather than ERA5 makes HRES outperform MLWP models.

\section{Discussion and Conclusion}\label{sec:discus}
We introduce \textbf{PhysMetrics.Weather}: an open-source, unified framework for quantifying the physical consistency of MLWP models by assessing (1) conservation laws, (2) spectral energy distribution, and (3) dynamical balances. This is an important step forward for the MLWP community which has so far primarily relied on metrics for predictive skill or accuracy.

Moreover, we observe that models optimized strictly for point-wise error (e.g., MSE, MAE) exhibit a tendency toward physical drift, including violations of energy and mass conservation, reductions in effective resolution (spatial smoothing), and unrealistic deviations from equilibria such as the hydrostatic balance. Consistent with prior work \citep{lam_graphcast_2023,hoffman_distortion_1995,subich_fixing_2025}, PhysMetrics.Weather confirms and quantifies that deterministic MLWP models are prone to smoothing atmospheric features to minimize average errors over time. Furthermore, it shows the value of hybrid architectures: by coupling neural networks with classical, differentiable physics solvers, models like NeuralGCM maintain stability across all evaluated physical consistency metrics over extended prediction time periods. However, this comes at a development cost: NeuralGCM is approximately 6 times more computationally expensive to train than purely data-driven MLWPs like GraphCast or Pangu-Weather \citep{couairon2024archesweatherefficientaiweather}.

\textbf{Limitations.} We acknowledge several limitations of the current version of PhysMetrics.Weather. First, its scope is limited to the variables available in the WeatherBench 2 archive. Because quantities like precipitation, evaporation, and 3D vertical velocity are not available for all MLWPs, we cannot evaluate the hydrological cycle or 3D momentum advection. Therefore, PhysMetrics.Weather serves as an initial, extensible baseline rather than a final product. Second, our evaluation currently relies on the ERA5 reanalysis as a proxy for the real atmospheric state. Since ERA5 assimilates observational data, it is not bias-free and may exhibit minor drifts in total atmospheric energy or mass.

\textbf{Future Directions.} PhysMetrics.Weather's open-source design facilitates the easy integration of emerging variables, such as 3D vertical velocity, cloud liquid water, precipitation, and radiative fluxes. This will allow future versions to check for a closed hydrological budget, advective transport, or top-of-atmosphere energy balance, moving beyond large-scale dynamics to also test thermodynamic and climate-relevant constraints. Moreover, as the field pivots toward generative diffusion models (e.g., GenCast \citep{price_gencast_2024}), our evaluation must evolve. Since these models sample specific, highly detailed atmospheric states rather than a single, often smoothed deterministic average, future versions of PhysMetrics.Weather will be extended to evaluate probabilistic ensembles. By establishing this unified evaluation framework, PhysMetrics.Weather provides a set of metrics needed to determine model suitability for specific operational tasks, guide physics-informed architectures, and encourage the development of physically consistent weather forecast models.

\bibliographystyle{unsrtnat}
\bibliography{bibentries}

@article{bodnar_foundation_2025,
	title = {A foundation model for the {Earth} system},
	volume = {641},
	copyright = {2025 The Author(s)},
	issn = {1476-4687},
	url = {https://www.nature.com/articles/s41586-025-09005-y},
	doi = {10.1038/s41586-025-09005-y},
	language = {en},
	number = {8065},
	urldate = {2025-11-14},
	journal = {Nature},
	publisher = {Nature Publishing Group},
	author = {Bodnar, Cristian and Bruinsma, Wessel P. and Lucic, Ana and Stanley, Megan and Allen, Anna and Brandstetter, Johannes and Garvan, Patrick and Riechert, Maik and Weyn, Jonathan A. and Dong, Haiyu and Gupta, Jayesh K. and Thambiratnam, Kit and Archibald, Alexander T. and Wu, Chun-Chieh and Heider, Elizabeth and Welling, Max and Turner, Richard E. and Perdikaris, Paris},
	month = may,
	year = {2025},
	pages = {1180--1187},
}

@article{kochkov_neural_2024,
	title = {Neural general circulation models for weather and climate},
	volume = {632},
	copyright = {2024 The Author(s)},
	issn = {1476-4687},
	url = {https://www.nature.com/articles/s41586-024-07744-y},
	doi = {10.1038/s41586-024-07744-y},
	language = {en},
	number = {8027},
	urldate = {2025-11-14},
	journal = {Nature},
	publisher = {Nature Publishing Group},
	author = {Kochkov, Dmitrii and Yuval, Janni and Langmore, Ian and Norgaard, Peter and Smith, Jamie and Mooers, Griffin and Klöwer, Milan and Lottes, James and Rasp, Stephan and Düben, Peter and Hatfield, Sam and Battaglia, Peter and Sanchez-Gonzalez, Alvaro and Willson, Matthew and Brenner, Michael P. and Hoyer, Stephan},
	month = aug,
	year = {2024},
	pages = {1060--1066},
}

@misc{noaa2004gfs,
  author       = {{NOAA}},
  title        = {Global Forecast System (GFS) 1.0 Degree},
  year         = {2004},
  publisher    = {NOAA National Centers for Environmental Information},
  note         = {NCEI DSI 6174.},
  howpublished = {Dataset},
  url          = {https://www.ncdc.noaa.gov/data-access/model-data/model-datasets/global-forecast-system-gfs}
}

@book{ecmwf_ifs_dynamics_2024,
  author       = {{ECMWF}},
  title        = {IFS Documentation CY49R1 - Part III: Dynamics and Numerical Procedures},
  year         = {2024},
  publisher    = {ECMWF},
  url          = {https://www.ecmwf.int/en/elibrary/81625-ifs-documentation-cy49r1-part-iii-dynamics-and-numerical-procedures},
  doi          = {10.21957/d04fb7a27e}
}

@misc{subramaniam_imposing_2025,
	title = {Imposing the {Fundamental} {Dynamical} {Constraint} of {Hydrostatic} {Balance} to {Improve} {Global} {ML} {Weather} {Prediction}},
	url = {http://arxiv.org/abs/2506.08285},
	doi = {10.48550/arXiv.2506.08285},
	urldate = {2026-01-22},
	publisher = {arXiv},
	author = {Subramaniam, Akshay and Durran, Dale and Pruitt, David and Cresswell-Clay, Nathaniel and Yik, William},
	month = jun,
	year = {2025},
	note = {arXiv:2506.08285 [physics]},
}

@misc{bodnar_foundation_2024,
	title = {A {Foundation} {Model} for the {Earth} {System}},
	url = {http://arxiv.org/abs/2405.13063},
	doi = {10.48550/arXiv.2405.13063},
	urldate = {2026-01-22},
	publisher = {arXiv},
	author = {Bodnar, Cristian and Bruinsma, Wessel P. and Lucic, Ana and Stanley, Megan and Vaughan, Anna and Brandstetter, Johannes and Garvan, Patrick and Riechert, Maik and Weyn, Jonathan A. and Dong, Haiyu and Gupta, Jayesh K. and Thambiratnam, Kit and Archibald, Alexander T. and Wu, Chun-Chieh and Heider, Elizabeth and Welling, Max and Turner, Richard E. and Perdikaris, Paris},
	month = nov,
	year = {2024},
	note = {arXiv:2405.13063 [physics]},
}

@misc{lam_graphcast_2023,
	title = {{GraphCast}: {Learning} skillful medium-range global weather forecasting},
	shorttitle = {{GraphCast}},
	url = {http://arxiv.org/abs/2212.12794},
	doi = {10.48550/arXiv.2212.12794},
	urldate = {2026-01-22},
	publisher = {arXiv},
	author = {Lam, Remi and Sanchez-Gonzalez, Alvaro and Willson, Matthew and Wirnsberger, Peter and Fortunato, Meire and Alet, Ferran and Ravuri, Suman and Ewalds, Timo and Eaton-Rosen, Zach and Hu, Weihua and Merose, Alexander and Hoyer, Stephan and Holland, George and Vinyals, Oriol and Stott, Jacklynn and Pritzel, Alexander and Mohamed, Shakir and Battaglia, Peter},
	month = aug,
	year = {2023},
	note = {arXiv:2212.12794 [cs]},
}

@misc{pathak_fourcastnet_2022,
	title = {{FourCastNet}: {A} {Global} {Data}-driven {High}-resolution {Weather} {Model} using {Adaptive} {Fourier} {Neural} {Operators}},
	shorttitle = {{FourCastNet}},
	url = {http://arxiv.org/abs/2202.11214},
	doi = {10.48550/arXiv.2202.11214},
	urldate = {2026-01-22},
	publisher = {arXiv},
	author = {Pathak, Jaideep and Subramanian, Shashank and Harrington, Peter and Raja, Sanjeev and Chattopadhyay, Ashesh and Mardani, Morteza and Kurth, Thorsten and Hall, David and Li, Zongyi and Azizzadenesheli, Kamyar and Hassanzadeh, Pedram and Kashinath, Karthik and Anandkumar, Animashree},
	month = feb,
	year = {2022},
	note = {arXiv:2202.11214 [physics]},
}

@misc{bi_pangu-weather_2022,
	title = {Pangu-{Weather}: {A} {3D} {High}-{Resolution} {Model} for {Fast} and {Accurate} {Global} {Weather} {Forecast}},
	shorttitle = {Pangu-{Weather}},
	url = {http://arxiv.org/abs/2211.02556},
	doi = {10.48550/arXiv.2211.02556},
	urldate = {2026-01-22},
	publisher = {arXiv},
	author = {Bi, Kaifeng and Xie, Lingxi and Zhang, Hengheng and Chen, Xin and Gu, Xiaotao and Tian, Qi},
	month = nov,
	year = {2022},
	note = {arXiv:2211.02556 [physics]},
}

@misc{verma_climode_2024,
	title = {{ClimODE}: {Climate} and {Weather} {Forecasting} with {Physics}-informed {Neural} {ODEs}},
	shorttitle = {{ClimODE}},
	url = {http://arxiv.org/abs/2404.10024},
	doi = {10.48550/arXiv.2404.10024},
	urldate = {2026-01-22},
	publisher = {arXiv},
	author = {Verma, Yogesh and Heinonen, Markus and Garg, Vikas},
	month = apr,
	year = {2024},
	note = {arXiv:2404.10024 [cs]},
}

@article{hoffman_distortion_1995,
	chapter = {Monthly Weather Review},
	title = {Distortion {Representation} of {Forecast} {Errors}},
	volume = {123},
	issn = {1520-0493, 0027-0644},
	url = {https://journals.ametsoc.org/view/journals/mwre/123/9/1520-0493_1995_123_2758_drofe_2_0_co_2.xml},
	doi = {10.1175/1520-0493(1995)123<2758:DROFE>2.0.CO;2},
	language = {EN},
	number = {9},
	urldate = {2026-01-22},
	journal = {Monthly Weather Review},
	publisher = {American Meteorological Society},
	author = {Hoffman, Ross N. and Liu, Zheng and Louis, Jean-Francois and Grassoti, Christopher},
	month = sep,
	year = {1995},
	pages = {2758--2770},
}

@misc{rasp_weatherbench_2024,
	title = {{WeatherBench} 2: {A} benchmark for the next generation of data-driven global weather models},
	shorttitle = {{WeatherBench} 2},
	url = {http://arxiv.org/abs/2308.15560},
	doi = {10.48550/arXiv.2308.15560},
	urldate = {2026-01-22},
	publisher = {arXiv},
	author = {Rasp, Stephan and Hoyer, Stephan and Merose, Alexander and Langmore, Ian and Battaglia, Peter and Russel, Tyler and Sanchez-Gonzalez, Alvaro and Yang, Vivian and Carver, Rob and Agrawal, Shreya and Chantry, Matthew and Bouallegue, Zied Ben and Dueben, Peter and Bromberg, Carla and Sisk, Jared and Barrington, Luke and Bell, Aaron and Sha, Fei},
	month = jan,
	year = {2024},
	note = {arXiv:2308.15560 [physics]},
}

@article{bonavita_limitations_2024,
	title = {On {Some} {Limitations} of {Current} {Machine} {Learning} {Weather} {Prediction} {Models}},
	volume = {51},
	copyright = {© 2024 ECMWF.},
	issn = {1944-8007},
	url = {https://onlinelibrary.wiley.com/doi/abs/10.1029/2023GL107377},
	doi = {10.1029/2023GL107377},
	language = {en},
	number = {12},
	urldate = {2026-01-22},
	journal = {Geophysical Research Letters},
	author = {Bonavita, Massimo},
	year = {2024},
	note = {\_eprint: https://agupubs.onlinelibrary.wiley.com/doi/pdf/10.1029/2023GL107377},
	pages = {e2023GL107377},
}

@misc{sha_improving_2025-1,
	title = {Improving {AI} weather prediction models using global mass and energy conservation schemes},
	url = {http://arxiv.org/abs/2501.05648},
	doi = {10.48550/arXiv.2501.05648},
	urldate = {2026-01-22},
	publisher = {arXiv},
	author = {Sha, Yingkai and Schreck, John S. and Chapman, William and Gagne, David John},
	month = jan,
	year = {2025},
	note = {arXiv:2501.05648 [physics]},
}

@misc{dunstan_fastnet_2025,
	title = {{FastNet}: {Improving} the physical consistency of machine-learning weather prediction models through loss function design},
	shorttitle = {{FastNet}},
	url = {http://arxiv.org/abs/2509.17601},
	doi = {10.48550/arXiv.2509.17601},
	urldate = {2026-01-22},
	publisher = {arXiv},
	author = {Dunstan, Tom and Strickson, Oliver and Bennett, Thusal and Bowyer, Jack and Burnand, Matthew and Chappell, James and Coca-Castro, Alejandro and Dale, Kirstine Ida and Daub, Eric G. and Eftekhari, Noushin and Janmaijaya, Manvendra and Lillis, Jon and Salvador-Jasin, David and Simpson, Nathan and Chan, Ryan Sze-Yin and Elmasri, Mohamad and France, Lydia Allegranza and Madge, Sam and Bokeria, Levan and Brown, Hannah and Dodds, Tom and Ellis, Anna-Louise and Llewellyn-Jones, David and McCaie, Theo and Moreton, Sophia and Potter, Tom and Robinson, James and Scaife, Adam A. and Stenson, Iain and Walters, David and Bett-Williams, Karina and Zeeland, Louisa van and Yatsyshin, Peter and Hosking, J. Scott},
	month = sep,
	year = {2025},
	note = {arXiv:2509.17601 [physics]},
}

@misc{saccardi_assessing_2025,
	title = {Assessing the {Geographic} {Generalization} and {Physical} {Consistency} of {Generative} {Models} for {Climate} {Downscaling}},
	url = {http://arxiv.org/abs/2510.13722},
	doi = {10.48550/arXiv.2510.13722},
	urldate = {2026-01-22},
	publisher = {arXiv},
	author = {Saccardi, Carlo and Pierzyna, Maximilian and Borde, Haitz Sáez de Ocáriz and Monaco, Simone and Meo, Cristian and Liò, Pietro and Saathof, Rudolf and Joseph, Geethu and Dauwels, Justin},
	month = oct,
	year = {2025},
	note = {arXiv:2510.13722 [cs]},
}

@misc{subich_fixing_2025,
	title = {Fixing the {Double} {Penalty} in {Data}-{Driven} {Weather} {Forecasting} {Through} a {Modified} {Spherical} {Harmonic} {Loss} {Function}},
	url = {http://arxiv.org/abs/2501.19374},
	doi = {10.48550/arXiv.2501.19374},
	urldate = {2026-01-22},
	publisher = {arXiv},
	author = {Subich, Christopher and Husain, Syed Zahid and Separovic, Leo and Yang, Jing},
	month = may,
	year = {2025},
	note = {arXiv:2501.19374 [cs]},
}

@misc{nathaniel_chaosbench_2024,
	title = {{ChaosBench}: {A} {Multi}-{Channel}, {Physics}-{Based} {Benchmark} for {Subseasonal}-to-{Seasonal} {Climate} {Prediction}},
	shorttitle = {{ChaosBench}},
	url = {http://arxiv.org/abs/2402.00712},
	doi = {10.48550/arXiv.2402.00712},
	urldate = {2026-01-22},
	publisher = {arXiv},
	author = {Nathaniel, Juan and Qu, Yongquan and Nguyen, Tung and Yu, Sungduk and Busecke, Julius and Grover, Aditya and Gentine, Pierre},
	month = nov,
	year = {2024},
	note = {arXiv:2402.00712 [cs]},
}

@misc{noauthor_brightbandtechextremeweatherbench_2026,
	title = {brightbandtech/{ExtremeWeatherBench}},
	copyright = {MIT},
    author = {Brightband},
	url = {https://github.com/brightbandtech/ExtremeWeatherBench},
	urldate = {2026-01-23},
	publisher = {Brightband},
	month = jan,
	year = {2026},
	note = {original-date: 2024-08-15T15:33:50Z},
}

@misc{sun_fuxi_2024,
	title = {{FuXi} {Weather}: {A} data-to-forecast machine learning system for global weather},
	shorttitle = {{FuXi} {Weather}},
	url = {http://arxiv.org/abs/2408.05472},
	doi = {10.48550/arXiv.2408.05472},
	urldate = {2026-01-29},
	publisher = {arXiv},
	author = {Sun, Xiuyu and Zhong, Xiaohui and Xu, Xiaoze and Huang, Yuanqing and Li, Hao and Neelin, J. David and Chen, Deliang and Feng, Jie and Han, Wei and Wu, Libo and Qi, Yuan},
	month = nov,
	year = {2024},
	note = {arXiv:2408.05472 [cs]},
}

@misc{price_gencast_2024,
	title = {{GenCast}: {Diffusion}-based ensemble forecasting for medium-range weather},
	shorttitle = {{GenCast}},
	url = {http://arxiv.org/abs/2312.15796},
	doi = {10.48550/arXiv.2312.15796},
	urldate = {2026-01-29},
	publisher = {arXiv},
	author = {Price, Ilan and Sanchez-Gonzalez, Alvaro and Alet, Ferran and Andersson, Tom R. and El-Kadi, Andrew and Masters, Dominic and Ewalds, Timo and Stott, Jacklynn and Mohamed, Shakir and Battaglia, Peter and Lam, Remi and Willson, Matthew},
	month = may,
	year = {2024},
	note = {arXiv:2312.15796 [cs]},
}

@article{nastrom1985climatology,
  title={A climatology of atmospheric wavenumber spectra of wind and temperature observed by commercial aircraft},
  author={Nastrom, G. D. and Gage, K. S.},
  journal={Journal of the Atmospheric Sciences},
  volume={42},
  number={9},
  pages={950--960},
  year={1985},
  doi={10.1175/1520-0469(1985)042<0950:ACOAWS>2.0.CO;2}
}

@misc{Hersbach2023ERA5,
  author       = {Hersbach, H. and Bell, B. and Berrisford, P. and Biavati, G. and Hor\'anyi, A. and Mu\~noz Sabater, J. and Nicolas, J. and Peubey, C. and Radu, R. and Rozum, I. and Schepers, D. and Simmons, A. and Soci, C. and Dee, D. and Th\'epaut, J.-N.},
  title        = {{ERA5 hourly data on single levels from 1940 to present}},
  year         = {2023},
  howpublished = {Copernicus Climate Change Service (C3S) Climate Data Store (CDS)},
  doi          = {10.24381/cds.adbb2d47},
  url          = {https://doi.org/10.24381/cds.adbb2d47}
}

@misc{ecmwf_hres_2024,
  author       = {{ECMWF}},
  title        = {Atmospheric Model High Resolution Forecast ({Set I - HRES})},
  year         = {2024},
  url          = {https://www.ecmwf.int/en/forecasts/datasets/set-i},
  note         = {Accessed: 2024},
  howpublished = {ECMWF Forecasts Dataset Documentation}
}

@article{rasp_weatherbench_2020,
	title = {{WeatherBench}: {A} {Benchmark} {Data} {Set} for {Data}-{Driven} {Weather} {Forecasting}},
	volume = {12},
	issn = {1942-2466},
	shorttitle = {{WeatherBench}},
	url = {https://agupubs.onlinelibrary.wiley.com/doi/10.1029/2020MS002203},
	doi = {10.1029/2020MS002203},
	language = {en},
	number = {11},
	urldate = {2026-02-10},
	journal = {Journal of Advances in Modeling Earth Systems},
	publisher = {John Wiley \& Sons, Ltd},
	author = {Rasp, Stephan and Dueben, Peter D. and Scher, Sebastian and Weyn, Jonathan A. and Mouatadid, Soukayna and Thuerey, Nils},
	month = nov,
	year = {2020},
	pages = {e2020MS002203},
}

@misc{nguyen2023climatelearnbenchmarkingmachinelearning,
      title={ClimateLearn: Benchmarking Machine Learning for Weather and Climate Modeling}, 
      author={Tung Nguyen and Jason Jewik and Hritik Bansal and Prakhar Sharma and Aditya Grover},
      year={2023},
      eprint={2307.01909},
      archivePrefix={arXiv},
      primaryClass={cs.LG},
      url={https://arxiv.org/abs/2307.01909}, 
}

@misc{cmip6_highresmip_2018,
  title = {{CMCC CMCC-CM2-VHR4} model output prepared for {CMIP6 HighResMIP} hist-1950},
  author = {Scoccimarro, Enrico and Bellucci, Alessio and Peano, Daniele},
  publisher = {Earth System Grid Federation},
  year = {2018},
  doi = {10.22033/ESGF/CMIP6.3818},
  howpublished = {\url{https://doi.org/10.22033/ESGF/CMIP6.3818}}
}

@misc{weatherbench_x_2025,
  author = {Rasp, Stephan and others},
  title = {WeatherBench-X: A modular framework for evaluating weather forecasts},
  year = {2025},
  publisher = {GitHub},
  journal = {GitHub repository},
  howpublished = {\url{https://github.com/google-research/weatherbenchX}}
}

@article{reichstein2019deep,
  title={Deep learning and process understanding for data-driven Earth system science},
  author={Reichstein, Markus and Camps-Valls, Gustau and Stevens, Bjorn and Jung, Martin and Denzler, Joachim and Carvalhais, Nuno and Prabhat},
  journal={Nature},
  volume={566},
  number={7743},
  pages={195--204},
  year={2019},
  publisher={Nature Publishing Group UK London},
  doi={10.1038/s41586-019-0912-1}
}

@article{PhysRevLett.126.098302,
  title = {Enforcing Analytic Constraints in Neural Networks Emulating Physical Systems},
  author = {Beucler, Tom and Pritchard, Michael and Rasp, Stephan and Ott, Jordan and Baldi, Pierre and Gentine, Pierre},
  journal = {Phys. Rev. Lett.},
  volume = {126},
  issue = {9},
  pages = {098302},
  numpages = {7},
  year = {2021},
  month = {Mar},
  publisher = {American Physical Society},
  doi = {10.1103/PhysRevLett.126.098302},
  url = {https://link.aps.org/doi/10.1103/PhysRevLett.126.098302}
}

@article{schultz2021can,
  title={Can deep learning beat numerical weather prediction?},
  author={Schultz, Martin G and Betancourt, Clara and Gong, Bing and Kleinert, Felix and Langguth, Michael and Leufen, Lukas Hubert and Mozaffari, Amirpasha and Stadtler, Scarlet},
  journal={Philosophical Transactions of the Royal Society A: Mathematical, Physical and Engineering Sciences},
  volume={379},
  number={2194},
  year={2021},
  publisher={The Royal Society}
}

@article{fortin2014why,
  title={Why should ensemble spread match the {RMSE} of the ensemble mean?},
  author={Fortin, Vincent and Abaza, Mohamed and Anctil, Fran{\c{c}}ois and Turcotte, Richard},
  journal={Journal of Hydrometeorology},
  volume={15},
  number={4},
  pages={1708--1713},
  year={2014},
  doi={10.1175/JHM-D-14-0008.1}
}

@article{gneiting2007strictly,
  title={Strictly proper scoring rules, prediction, and estimation},
  author={Gneiting, Tilmann and Raftery, Adrian E},
  journal={Journal of the American Statistical Association},
  volume={102},
  number={477},
  pages={359--378},
  year={2007},
  publisher={Taylor \& Francis},
  doi={10.1198/016214506000001437}
}

@book{holton2012dynamic,
  title={An Introduction to Dynamic Meteorology},
  author={Holton, James R and Hakim, Gregory J},
  edition={5th},
  year={2012},
  publisher={Academic press}
}

@article{trenberth2005mass,
  title={The mass of the atmosphere: A constraint on global analyses},
  author={Trenberth, Kevin E and Smith, Lesley},
  journal={Journal of Climate},
  volume={18},
  number={6},
  pages={864--875},
  year={2005}
}

@article{NorthernHemisphereMidlatitudeCycloneIntensityBiasesinMachineLearningWeatherPredictionModels,
  author  = {Helen F. Dacre and Andrew J. Charlton-Perez and Simon Driscoll and Sue L. Gray and Ben Harvey and Natalie J. Harvey and Kevin I. Hodges and Kieran M. R. Hunt and Ambrogio Volont{\`e}},
  title   = {Northern Hemisphere Midlatitude Cyclone Intensity Biases in Machine Learning Weather Prediction Models},
  journal = {Bulletin of the American Meteorological Society},
  year    = {2026},
  volume  = {107},
  number  = {1},
  pages   = {E208--E221},
  doi     = {10.1175/BAMS-D-25-0129.1},
  url     = {https://journals.ametsoc.org/view/journals/bams/107/1/BAMS-D-25-0129.1.xml}
}

@article{pappenberger2024machine,
  title={Machine learning and physics in weather forecasting: a discussion between Alan Thorpe and Florian Pappenberger},
  author={Pappenberger, Florian and Thorpe, Alan},
  journal={ECMWF -- In Focus},
  year={2024},
  month={jun},
  url={https://www.ecmwf.int/en/about/media-centre/focus/2024/machine-learning-and-physics-weather-forecasting-discussion-0}
}

@article{fan2025incorporating,
  title={Incorporating Multivariate Consistency in ML-Based Weather Forecasting with Latent-space Constraints},
  author={Fan, Hang and Xiao, Yi and Qu, Yongquan and Ling, Fenghua and Fei, Ben and Bai, Lei and Gentine, Pierre},
  journal={arXiv preprint arXiv:2510.04006},
  year={2025},
  url={https://arxiv.org/abs/2510.04006}
}

@article{Stone_and_Carlson_1979,
    author = "Peter H.  Stone and John H.  Carlson",
    title = "Atmospheric Lapse Rate Regimes and Their Parameterization",
    journal = "Journal of Atmospheric Sciences",
    year = "1979",
    publisher = "American Meteorological Society",
    address = "Boston MA, USA",
    volume = "36",
    number = "3",
    doi = "10.1175/1520-0469(1979)036<0415:ALRRAT>2.0.CO;2",
    pages=      "415 - 423",
    url = "https://journals.ametsoc.org/view/journals/atsc/36/3/1520-0469_1979_036_0415_alrrat_2_0_co_2.xml"
}

@misc{couairon2024archesweatherefficientaiweather,
      title={ArchesWeather: An efficient AI weather forecasting model at 1.5° resolution}, 
      author={Guillaume Couairon and Christian Lessig and Anastase Charantonis and Claire Monteleoni},
      year={2024},
      eprint={2405.14527},
      archivePrefix={arXiv},
      primaryClass={cs.LG},
      url={https://arxiv.org/abs/2405.14527}, 
}

@book{us_standard_atmosphere_1976,
  title={U.S. Standard Atmosphere, 1976},
  author={{NOAA} and {NASA} and {USAF}},
  year={1976},
  publisher={U.S. Government Printing Office},
  address={Washington, D.C.}
}
\appendix

\newpage
\section{Appendix A: Granular Preprocessing and Integration Formalisms}\label{app:integration}
 \subsection{Boundary Derivation Strategies}
 To compute the vertical column integrals detailed in the main text, a lower boundary (surface pressure, $p_s$) is required. Because MLWP models exhibit varying degrees of available output, we use a derivation strategy:
 
 \textbf{Direct output:} If surface pressure is directly provided by the model (e.g., IFS HRES), it is used without modification.
 
\textbf{Standard Atmosphere Derivation:} For models providing only mean sea level pressure ($p_\text{MSL}$) and no explicit surface pressure, $p_s$ is derived using the U.S. Standard Atmosphere temperature profile \citep{us_standard_atmosphere_1976}. By converting the surface geopotential ($\Phi_s$) to geometric height ($z = \Phi_s / g$), the surface pressure is estimated as:
$$p_s = p_\text{MSL} \cdot \left( 1 - \frac{\Gamma_\text{std} \cdot z}{T_0} \right)^{\frac{g}{R_d \Gamma_\text{std}}}$$
where $T_0 = 288.15 \text{ K}$ is the standard sea-level temperature, and $\Gamma_\text{std} = 0.0065 \text{ K/m}$ is the standard tropospheric lapse rate.
 
\textbf{External Surface Pressure Fallback:} For models where the WeatherBench 2 archive lacks both surface pressure ($p_s$) and mean sea-level pressure ($p_\text{MSL}$) (e.g., NeuralGCM), we substitute the reference (ERA5) surface pressure at the corresponding valid forecast time. We acknowledge that this substitution is imperfect. Because surface pressure is highly sensitive to orographic elevation, applying reference $p_s$ to a model operating at a different spatial resolution introduces structural inconsistencies due to mismatched topography. Furthermore, by supplying an external boundary condition to close the vertical integrals, the evaluation for these specific models ceases to be a strictly closed-system measurement of the model's internal mass and energy budgets. Nevertheless, as demonstrated in our ablation study (Appendix \ref{app:sp}), the long-term trends in our conservation metrics remain robust regardless of the surface pressure boundary condition used.
 
\subsection{Spatial Alignment and Grid Computations}\label{app:area_weights}

To perform the Spherical Harmonic Transform (SHT), the prediction and reference tensors must have identical spatial dimensions. A standard $0.25^\circ$ geographic grid spanning from $90^\circ$N to $90^\circ$S consists of 721 latitude points, explicitly including both poles. However, MLWP models typically output 720 latitude points to maintain even tensor dimensions for neural network padding and pooling. Consequently, the grids are identical except for a single unshared polar row. To achieve alignment, we simply drop this unshared row from the reference tensor. We strictly avoid regridding or interpolating the data, as spatial interpolation inherently acts as a low-pass filter, which would artificially alter the spectral energy distribution and corrupt the mass conservation budgets we intend to evaluate.

\begin{figure}
    \centering
    \includegraphics[width=0.9\linewidth]{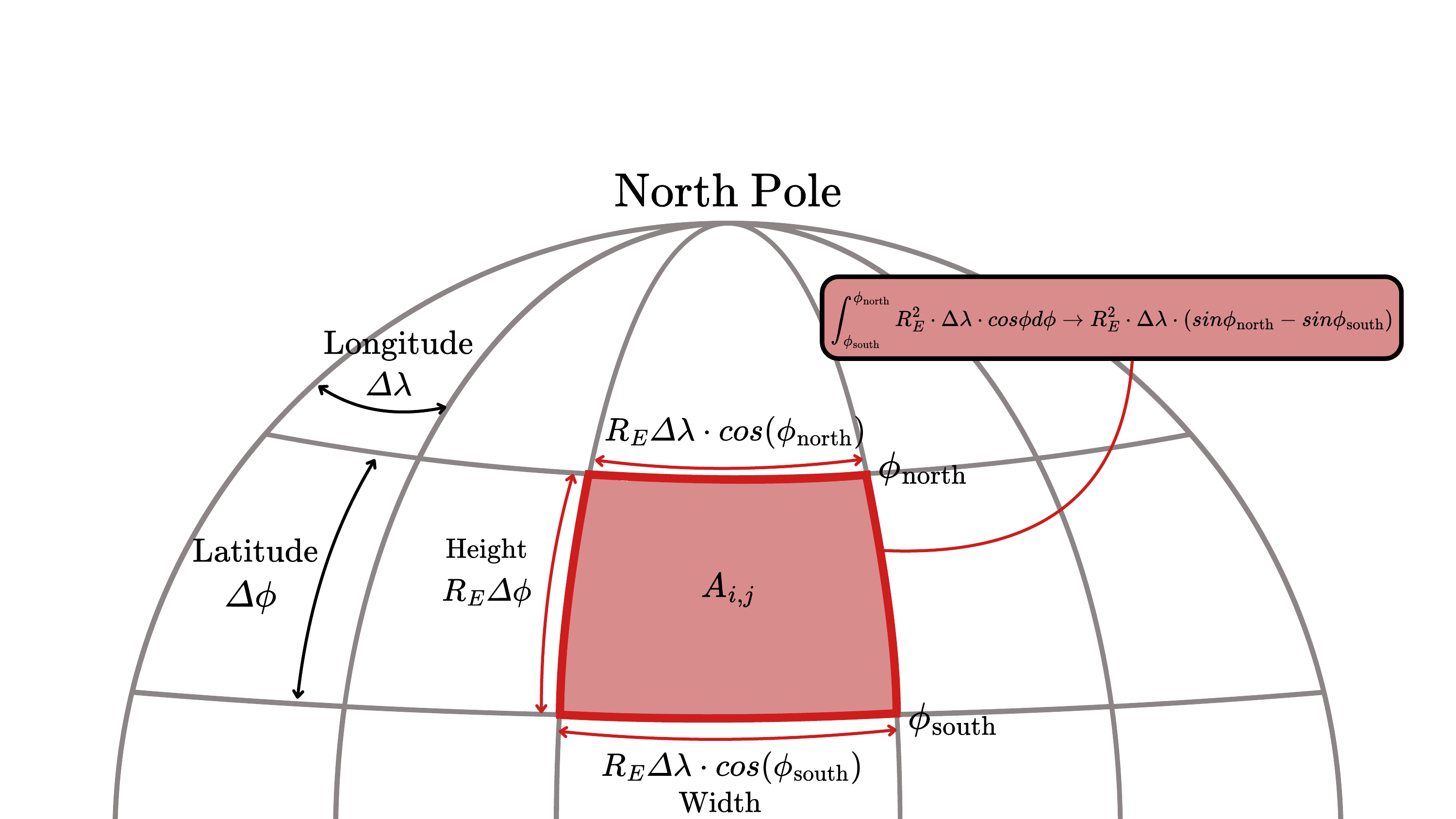}
    \caption{Geometric derivation of the grid-cell area ($A_{i,j}$) on a spherical Earth. Because the physical width of a cell ($R_E\Delta\lambda \cdot \cos\phi$) shrinks as it approaches the poles, a simple rectangular approximation overestimates polar areas. The true physical area is computed by integrating the differential element across the exact latitudinal bounds $[\phi_{south}, \phi_{north}]$.}
    \label{fig:area}
\end{figure}

For global aggregations, each grid cell must be weighted by its true physical surface area, $A_{i,j}$. Because explicit spatial bound arrays are not consistently provided across the WeatherBench 2 Zarr stores, we derive the exact cell boundaries from the 1D coordinate arrays. The area is obtained by integrating the spherical surface area across the exact latitudinal boundaries $[\phi_{south}, \phi_{north}]$ of the cell (see Figure \ref{fig:area}):

$$A_{i,j} = R_E^2 \cdot \Delta \lambda \cdot \left( \sin(\phi_{north}) - \sin(\phi_{south}) \right)$$

Depending on the model, spatial grids are either node-centered (coordinate points fall exactly on the poles) or cell-centered. If a grid is node-centered, assuming a constant latitudinal width (e.g., $0.25^\circ$) for every cell introduces a geometric impossibility: a cell centered exactly at $90^\circ$N would theoretically extend to $90.125^\circ$N, geographically exceeding the pole. To prevent this, we calculate the exact boundaries of each cell by finding the mathematical midpoint between adjacent coordinate points, strictly clipping the outermost boundaries at $+90^\circ$ and $-90^\circ$.
 \subsection{Global and Discrete Vertical Integration}\label{app:integration_global}
  We define a discrete global integral operator, denoted by $\langle \cdot \rangle_\mathcal{D}$. While angled brackets are often used in computer science to denote expectation, here they represent the sum of all grid cells weighted by their surface area over the Earth domain $\mathcal{D}$. We adapt the integrals from \citet{sha_improving_2025-1} into the following discrete form:
\begin{equation*}
\langle X \rangle_\mathcal{D} = \sum_{i,j\in\mathcal{D}} A_{i,j} X_{i,j}.
\end{equation*}
where $X_{i,j}$ is the value of the physical field at grid cell $(i,j)$, and $A_{i,j}$ ensures each cell contributes proportionally to its surface area.

For vertical integration across $N+1$ discrete pressure levels $\{p_n\}_{n=0}^N$, the effective pressure thickness $\Delta p_n^*$ used in the trapezoidal integration scheme is defined as:
 
 $$\Delta p_n^* = \max\left(0, \min(p_{n+1}, p_s) - \min(p_n, p_s) \right)$$
 
 This ensures layers entirely above the surface are counted fully, while levels extending below the surface ($p_n \ge p_s$) are assigned zero thickness, preventing the calculation of non-existing atmospheric mass. Column-integrated quantities are then computed as:
\begin{equation*}
    \int_0^{p_s} X(p) \, dp \approx X_0 p_0 + \sum_{n=0}^{N-1} \frac{X_n + X_{n+1}}{2} \Delta p_n^* + X_N \max(0, p_s - p_N)
\end{equation*}

Here, the first term accounts for the mass of air from the top of the atmosphere (0 hPa) down to the lowest available pressure level $p_0$. The summation handles the interior layers via trapezoidal averaging. The final term extrapolates the lowest model level field $X_N$ down to the surface in regions where the surface pressure exceeds the lowest available pressure level.

\paragraph{A Note on the Hydrostatic Assumption in Mass Integration}
The derivation of total atmospheric mass via $p_s / g$ uses the hydrostatic approximation ($\partial p / \partial z = -\rho g$). If an MLWP model exhibits severe physical drift and violates the hydrostatic balance, surface pressure will no longer represent the gravitational weight of the air column above. In such cases the dry air mass metric may reflect both actual mass loss and the breakdown of the hydrostatic assumption.

\subsection{Area-Weighted Error Metrics}
To evaluate spatial fields on a spherical grid, point-wise errors must be scaled by the grid-cell surface area $A_{i,j}$. For a given residual field $R_{i,j}$ evaluated over a specific domain $\mathcal{D}_\text{valid}$, the area-weighted RMSE is formulated as:
$$\text{RMSE} = \sqrt{\frac{\sum_{(i,j) \in \mathcal{D}_\text{valid}} A_{i,j} R_{i,j}^2}{\sum_{(i,j) \in \mathcal{D}_\text{valid}} A_{i,j}}}$$

\textbf{Geostrophic Imbalance:} The residual $R_{i,j}$ is the vector difference between the predicted horizontal wind $(u,v)$ and the derived geostrophic wind $(u_g, v_g)$, defined as $R_{i,j} = \sqrt{(u - u_g)^2 + (v - v_g)^2}$. The valid domain $\mathcal{D}_\text{valid}$ explicitly excludes the equatorial region ($|\phi| < 10^\circ$) and the poles ($|\phi| \ge 89.9^\circ$).

\textbf{Hydrostatic Imbalance:} The residual is the absolute error of the hypsometric equation, $R_{i,j} = \epsilon_h$, evaluated globally over all grid cells.

\subsection{Virtual Temperature Calculation}

For hydrostatic balance evaluations, the density effect of moisture is accounted for via the mean virtual temperature $T_v$ of the layer. Virtual temperature is the temperature at which a theoretical dry air parcel would have a total pressure and density equal to the moist parcel of air:

$$T_v = T(1 + 0.6078q)$$

where $q$ is the specific humidity, and the coefficient 0.6078 approximates $(R_v/R_d - 1)$, correcting for the density difference between moist and dry air.

\paragraph{Missing Variable Fallback:} Because models exhibit varying degrees of available output, specific humidity ($q$) is not always available in WeatherBench 2 (e.g., FuXi). For these models, we evaluate the hydrostatic balance using the available air temperature ($T_v \approx T$). To ensure a fair comparison, we apply this exact same approximation ($T_v \approx T$) to the reference dataset when calculating its baseline residual. While this ignores the small density effects of atmospheric moisture, it allows for a first evaluation of models not providing specific humidity (see Appendix \ref{app:Tv} for an ablation study justifying this approximation).

\subsection{Computing the Mean Lapse Rate Wasserstein Distance}\label{app:wasserstein}
In Section \ref{sec:balance}, we evaluate thermodynamic stability by comparing the distributions of the atmospheric lapse rate ($\Gamma$) across different climate zones. To do this, we calculate the 1-Wasserstein distance ($W_1$) between the predicted and reference lapse rate distributions.

Since the atmospheric data is evaluated on a spherical latitude-longitude grid, the physical size of the grid cells shrinks toward the poles. To ensure accurate representation, we use the true physical surface area of each grid cell ($A_{i,j}$) as a weight when calculating the regional distributions. Atmospheric stability varies across climate zones, meaning their underlying lapse rate distributions are different. To prevent distinct regional errors from canceling out in a single global distribution, the area-weighted 1-Wasserstein distance is calculated independently for three distinct regions: the Northern Hemisphere mid-latitudes ($30^\circ\text{N}$ to $60^\circ\text{N}$), the Tropics ($30^\circ\text{S}$ to $30^\circ\text{N}$), and the Southern Hemisphere mid-latitudes ($60^\circ\text{S}$ to $30^\circ\text{S}$). Finally, the \textit{Mean Lapse Rate Wasserstein} ($\overline{W}_1$) is the simple average of these three regional distances.

\section{Appendix B: Nomenclature and Metric Summaries}\label{app:constants}



\subsection{Physical Constants and Variables}
\begin{table}[H]
\centering
\caption{Nomenclature of physical constants, variables, and symbols used in PhysMetrics.Weather.}
\label{tab:nomenclature}
\renewcommand{\arraystretch}{1.1} 
\begin{tabular}{@{}lll@{}}
\toprule
\textbf{Symbol} & \textbf{Description (Value)} & \textbf{Units} \\
\midrule
\multicolumn{3}{l}{\textit{Physical Constants}} \\
$g$ & Gravitational acceleration ($9.80665$) & m\,s$^{-2}$ \\
$R_E$ & Earth radius ($6.371 \times 10^{6}$) & m \\
$c_{pd}$ & Specific heat capacity of dry air ($1004.64$) & J\,kg$^{-1}$\,K$^{-1}$ \\
$c_{pv}$ & Specific heat capacity of water vapor ($1810.0$) & J\,kg$^{-1}$\,K$^{-1}$ \\
$L_v$ & Latent heat of vaporization ($2.501 \times 10^{6}$) & J\,kg$^{-1}$ \\
$R_d$ & Gas constant for dry air ($287.05$) & J\,kg$^{-1}$\,K$^{-1}$ \\
$R_v$ & Gas constant for water vapor ($461.5$) & J\,kg$^{-1}$\,K$^{-1}$ \\
$\Gamma_{std}$ & Standard lapse rate ($0.0065$) & K\,m$^{-1}$ \\
$\Omega$ & Earth angular velocity ($7.2921 \times 10^{-5}$) & rad\,s$^{-1}$ \\
\midrule
\multicolumn{3}{l}{\textit{Grid and Domain}} \\
$\lambda, \phi$ & Longitude and Latitude & rad \\
$\Delta \lambda$ & Longitudinal grid spacing & rad \\
$A_{i,j}$ & Grid-cell area & m$^2$ \\
$\mathcal{D}$ & Global Earth domain & - \\
$\mathcal{L}$ & Set of vertical pressure levels & - \\
$p$ & Pressure & Pa \\
$p_s$ & Surface pressure & Pa \\
$\Delta p^*$ & Effective pressure layer thickness & Pa \\
$t_0$ &  Current state time & h \\
$\Delta t$& Time step &  h\\
$\mathcal{T}$& Prediction time period & h\\
\midrule
\multicolumn{3}{l}{\textit{Atmospheric State Variables}} \\
$u, v$ & Zonal and Meridional wind components & m\,s$^{-1}$ \\
$u_g, v_g$ & Geostrophic wind components & m\,s$^{-1}$ \\
$T$ & Temperature & K \\
$T_v$ & Virtual temperature & K \\
$\Phi$ & Geopotential & m$^2$\,s$^{-2}$ \\
$q$ & Specific humidity & kg\,kg$^{-1}$ \\
$\Gamma$ & Environmental lapse rate & K\,km$^{-1}$ \\
$\text{TCWV}$ & Total column water vapor & kg\,m$^{-2}$ \\
$c_p$ & Specific heat capacity of moist air & J\,kg$^{-1}$\,K$^{-1}$ \\
\midrule
\multicolumn{3}{l}{\textit{Metrics and Spectral Analysis}} \\
$M_d, M_w$ & Global dry air mass and water mass & kg \\
$\mathcal{E}$ & Global total energy & J \\
$\delta_d, \delta_w, \delta_\mathcal{E}$ & Relative and anomalous drift rates & \%\,day$^{-1}$ \\
$\epsilon_h$ & Hydrostatic residual & m$^2$\,s$^{-2}$ \\
$k, m$ & Total and Zonal wavenumbers & - \\
$\hat{u}, \hat{v}$ & Spectral coefficients of wind & - \\
$E(k)$ & Kinetic energy spectrum & m$^2$\,s$^{-2}$ \\
$R(k)$ & Spectral energy ratio & - \\
$P(k)$ & Normalized spectral distribution & - \\
$L_{eff}$ & Effective resolution & m \\
$f$ & Coriolis parameter & s$^{-1}$ \\
$\overline{W}_1$ & Mean Lapse Rate Wasserstein & - \\
\bottomrule
\end{tabular}
\end{table}
\subsection{Summary of Evaluated Metrics}
\begin{table}[H]
\centering
\caption{Summary of physical consistency metrics evaluated in this framework.}
\label{tab:metrics_summary}
\renewcommand{\arraystretch}{1.3} 
\begin{tabularx}{\textwidth}{@{}llX@{}} 
\toprule
\textbf{Category} & \textbf{Metric} & \textbf{What it Measures} \\ 
\midrule
\multirow{3}{*}{\shortstack[l]{\textbf{Conservation} \\ \textbf{\& Variability}}} 
& Dry Air Mass Drift ($\delta_d$) & Global dry air mass drift over time; detects unphysical sources or sinks of atmospheric matter. \\
& Anomaly Water Drift ($\delta_w$) & Deviation of the model's total water mass trend from the reference data; detects systematic over-moistening or over-drying. \\
& Anomaly Energy Drift ($\delta_\mathcal{E}$) & Discrepancy in the model's total energy trend compared to the reference data; detects unphysical energy sinks or sources. \\ 
\midrule
\multirow{3}{*}{\shortstack[l]{\textbf{Structural} \\ \textbf{(Spectral)}}} 
& Effective Resolution ($L_{eff}$) & The physical scale at which the model loses significant kinetic energy; quantifies forecast blurring. \\
& Spectral Divergence ($W_1(P_\text{ref},P_\text{pred})$) & 1-Wasserstein distance of the kinetic energy spectrum; summarizes how much the model spectrum deviates from the reference spectrum across spatial scales. \\
& Spectral Residual (SpecRes) & Logarithmic RMSE of the energy spectrum; evaluates absolute magnitude of errors equally across low and high frequencies. \\ 
\midrule
\multirow{3}{*}{\shortstack[l]{\textbf{Dynamical} \\ \textbf{Balance}}} 
& Hydrostatic RMSE ($\text{RMSE}_{h}$) & Residual of the hydrostatic balance; measures vertical physical consistency between the mass (geopotential) and temperature. \\
& Geostrophic RMSE ($\text{RMSE}_{g}$) & Residual of geostrophic balance; measures horizontal consistency between the momentum (wind) and mass (geopotential). \\
& Mean Lapse Rate Wasserstein ($\overline{W}_1$) & Measures the capability to reproduce realistic distributions of atmospheric temperature lapse rates across different regions; evaluates vertical thermodynamic stability. \\
\bottomrule
\end{tabularx}
\end{table}

\section{Appendix C: Extended Results and Ablation Studies}\label{app:ablation}

\subsection{PhysMetrics.Weather with IFS as reference dataset}\label{app:ifs}
\subsubsection{Conservation}
\begin{figure*}[h]
    \centering
    
    \begin{subfigure}{\textwidth}
        \centering
        \includegraphics[width=0.95\linewidth]{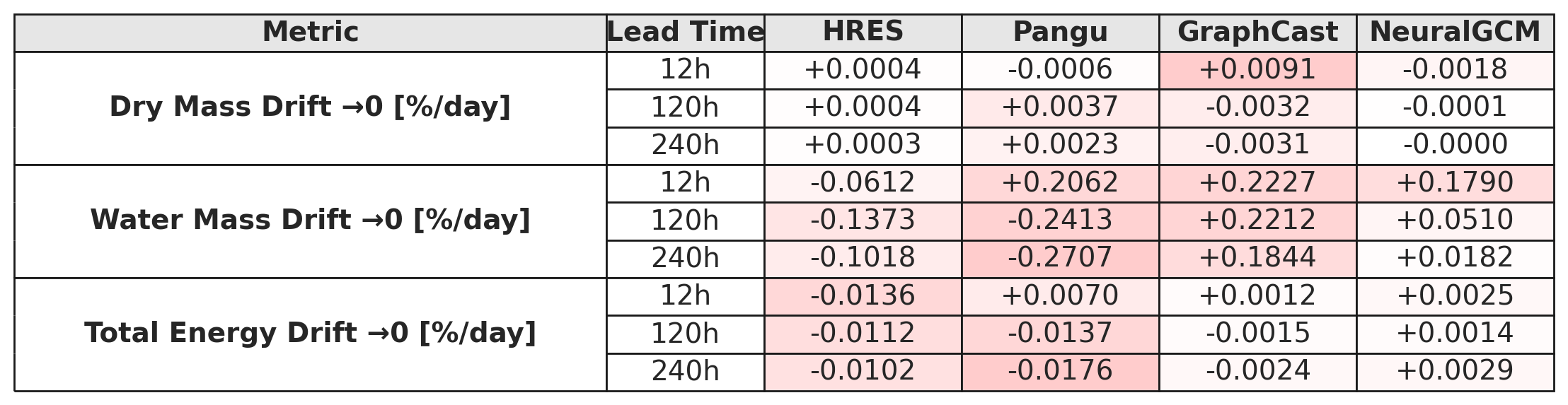} 
    \end{subfigure}

    \begin{subfigure}{\textwidth}
        \centering
        \includegraphics[width=0.95\linewidth]{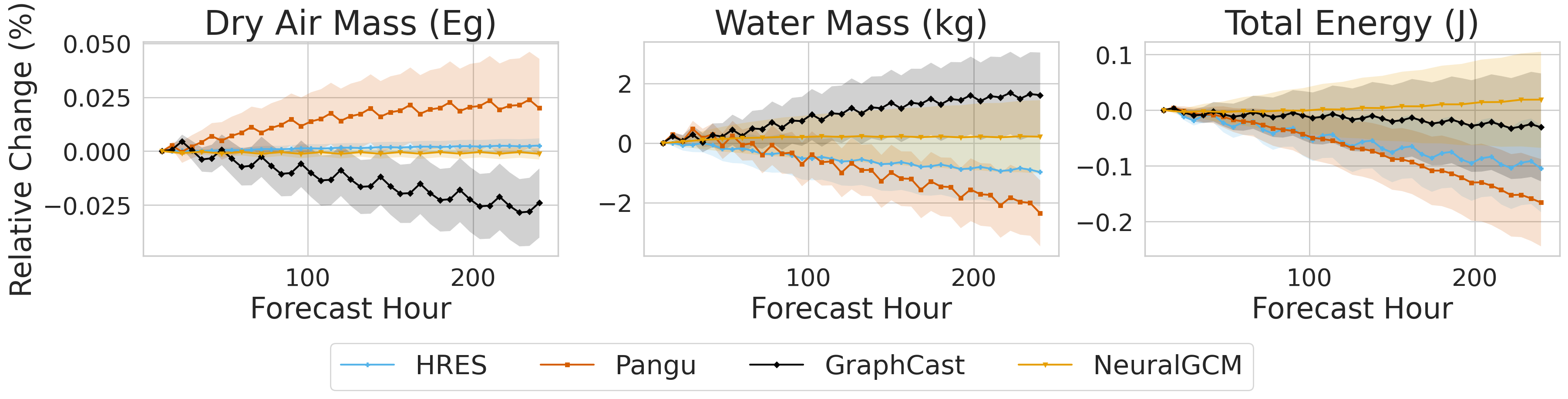}
    \end{subfigure}\hfill
    
    \caption{\textbf{Global budgets and time series for the conservation metrics over a 240-hour forecast period with an IFS HRES analysis as reference data.} \textbf{Top:} Summary of conservation metrics. \textbf{Bottom left:} Relative Dry Mass drift. \textbf{Bottom middle:} Anomaly Water Mass drift. \textbf{Bottom right:} Anomaly Total Energy drift. Constrained models like NeuralGCM maintain stable budgets over extended forecast periods, while MLWP models (e.g., GraphCast, Pangu-Weather) show divergence. Shading in time series indicates $\pm 1$ standard deviation over 2020.}
    \label{fig:conservation_metrics_IFS}
\end{figure*}

The time series presented in Figure \ref{fig:conservation_metrics_IFS} show that dry air mass, water mass, and total energy drift remain consistent with those derived using the ERA5 baseline (Figure \ref{fig:conservation_metrics}). While the summary metrics exhibit minor deviations, these differences primarily occur at a 12-h lead time. These short-term discrepancies are expected: because the drift metric calculates a linear trend over the forecast period, a short 12-h window is highly sensitive to the small differences in the initial atmospheric states between the ERA5 and IFS HRES datasets. Over extended lead times (e.g., 240 hours), the long-term trend dominates. This minimizes the impact of the initial dataset differences, yielding consistent overall drift rates regardless of which reference baseline is used.
\subsubsection{Energy Spectra}
\begin{figure*}[h]
    \centering
    
    \begin{subfigure}{\textwidth}
        \centering
        \includegraphics[width=0.93\linewidth]{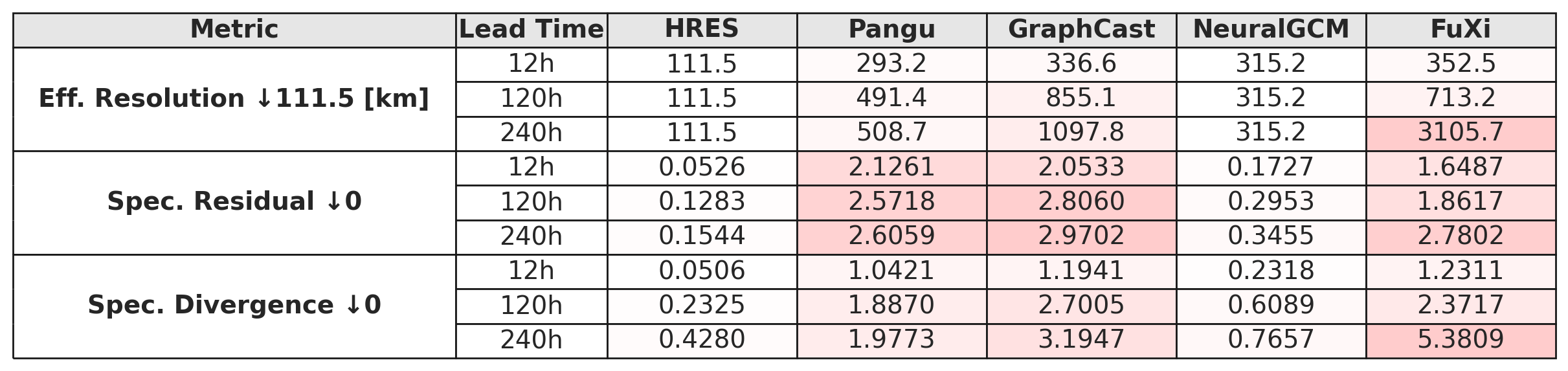} 
    \end{subfigure}

    \begin{subfigure}{\textwidth}
        \centering
        \includegraphics[width=0.93\linewidth]{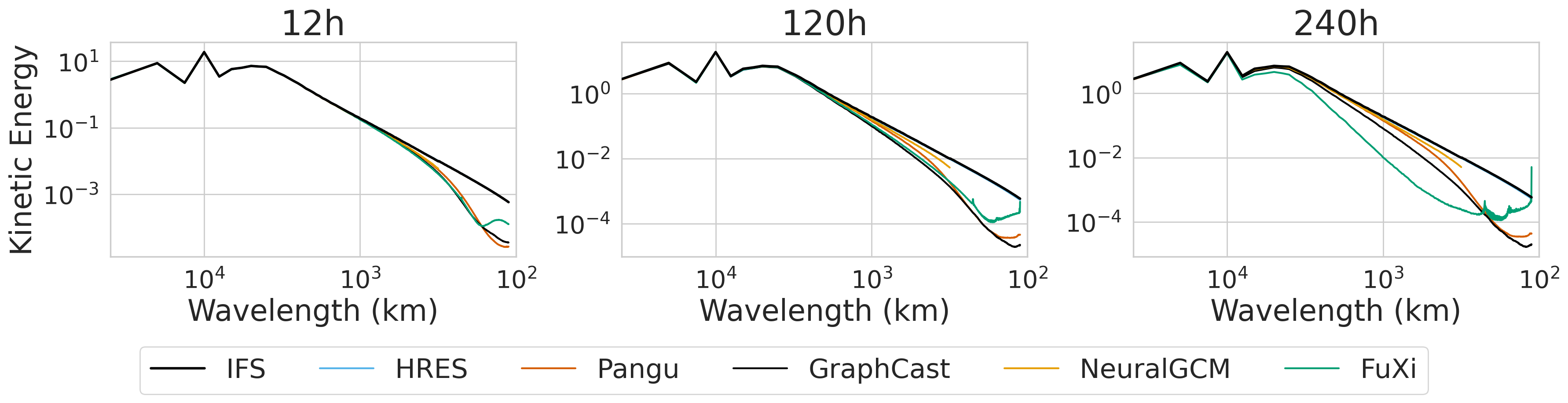}
    \end{subfigure}\hfill

    \caption{\textbf{KE spectra at 500 hPa with an IFS HRES analysis as reference.} \textbf{Top:} Summary of spectral metrics. \textbf{Bottom:} Kinetic energy spectra at 12-h, 120-h, and 240-h lead times, respectively.  Native resolutions: 0.25° (111.5 km), except NeuralGCM (0.7°, 315.2 km). Over extended forecasting periods, purely data-driven MLWP models exhibit spatial smoothing (loss of fine-scale energy) or artificial noise, whereas differentiable hybrid architectures preserve a realistic energy cascade.}
    \label{fig:combined_spectra_IFS}
\end{figure*}

Figure \ref{fig:combined_spectra_IFS} demonstrates that the IFS HRES analysis baseline includes more fine-scale KE than the ERA5 reanalysis. Evaluated against this new baseline, the IFS HRES operational forecasts align with the reference spectrum, eliminating the fine-scale "energy creation" artifact previously observed in Figure \ref{fig:combined_spectra}. This structural alignment is confirmed by the improved \textit{spectral residual} (SpecRes) and \textit{spectral divergence} (SpecDiv) scores for the HRES model.

The MLWP models perform worse on spectral metrics when evaluated against the HRES baseline. Because these data-driven models are trained to replicate ERA5, they naturally reproduce the smoother energy spectrum of their training data. For instance, GraphCast now yields a higher \textit{SpecRes} than FuXi because it cannot replicate the higher levels of fine-scale energy inherent to the operational HRES dataset. Despite these differences in absolute metric scores caused by switching the reference dataset, the overarching conclusion remains the same: MLWP models progressively lose effective resolution over extended forecast periods.
\subsubsection{Dynamical Balance}
\begin{figure*}[h]
    \centering
    
    \begin{subfigure}{\textwidth}
        \centering
        \includegraphics[width=0.95\linewidth]{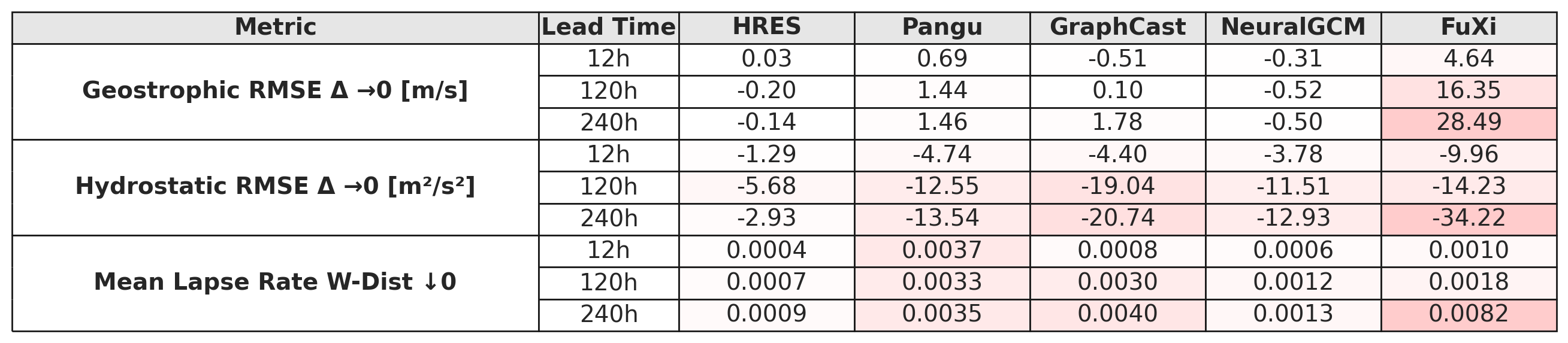} 
    \end{subfigure}

    \begin{subfigure}{\textwidth}
        \centering
        \includegraphics[width=0.95\linewidth]{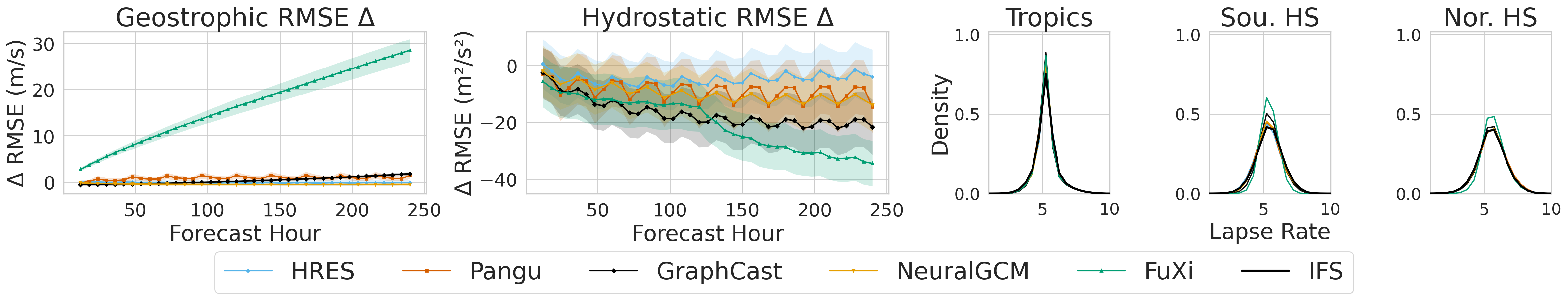}
    \end{subfigure}\hfill

    \caption{\textbf{Adherence to geostrophic wind balance, hydrostatic equilibrium and realistic atmospheric stability over a 240-hour horizon with an IFS HRES analysis as reference.} \textbf{Top:} Summary of dynamical balance metrics. \textbf{Bottom left:} Excess Geostrophic Imbalance. \textbf{Bottom middle:} Excess Hydrostatic Imbalance. \textbf{Bottom right:} Lapse rate distributions across different regions at 240-h lead time. Shading in time series indicates $\pm 1$ standard deviation over 2020.}
    \label{fig:RMSE_IFS}
\end{figure*}

Figure \ref{fig:RMSE_IFS} evaluates dynamical and thermodynamic equilibrium against the IFS HRES analysis. While the time series follow the same general trends as the ERA5 evaluation (Figure \ref{fig:RMSE}), the absolute magnitudes of the residuals shift uniformly. Because the reference dataset has changed, the \textit{excess hydrostatic imbalance} and \textit{excess geostrophic imbalance} show constant vertical offsets relative to the ERA5 results. However, this simply reflects the structural differences between the two reference baselines and does not alter the underlying conclusions regarding how the models perform over time.

The \textit{mean lapse rate Wasserstein} is also sensitive to the choice of reference data. HRES and GraphCast achieve improved thermodynamic stability scores when evaluated against the HRES baseline, whereas Pangu-Weather's performance degrades. FuXi initially demonstrates worse stability but converges to values comparable to its ERA5 evaluation by the 240-h lead time. Despite these numerical shifts in the Wasserstein distances, the overall shape, peak, and width of the lapse rate distributions (Figure \ref{fig:RMSE_IFS}d) strongly resemble the analysis using the ERA5 reference, confirming that the models' broad thermodynamic characteristics remain fundamentally the same.
\subsection{Sensitivity of Effective Resolution Parameters}\label{app:thresholding}
\begin{figure}[h]
    \centering
    \includegraphics[width=0.95\linewidth]{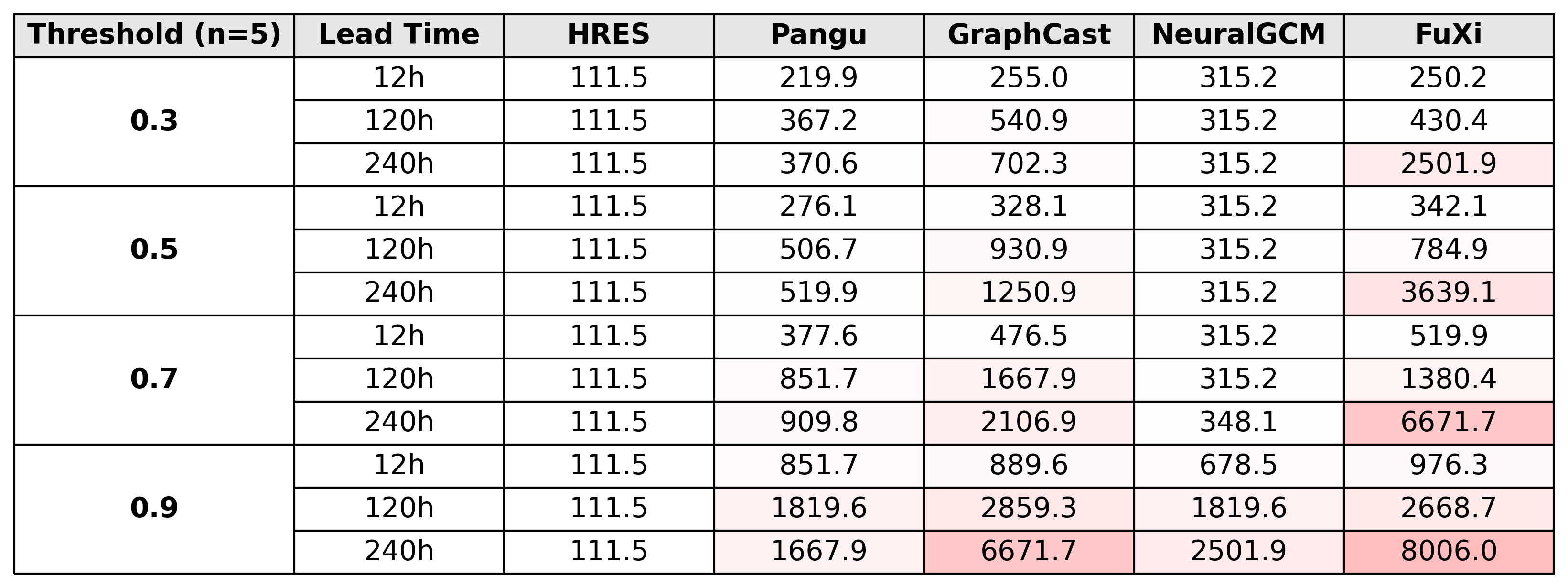}
    \caption{\textbf{Sensitivity of effective resolution to energy retention thresholds with ERA5 reference.} The effective resolution ($L_{eff}$, in km) of the 500 hPa KE spectrum for various models across 12-hour, 120-hour, and 240-hour lead times using retention thresholds of 0.3, 0.5, 0.7, and 0.9. Native resolutions: 0.25° (111.5 km), except NeuralGCM (0.7°, 315.2 km). Lower values indicate better retention of fine-scale physical features. A threshold of 0.5 is selected for the main evaluation as it optimally distinguishes the smoothing behaviors of MLWP models without over-penalizing minor spectral variations.}
    \label{fig:thresholding}
\end{figure}

Figure \ref{fig:thresholding} illustrates the sensitivity of the \textit{effective resolution} of the 500 hPa KE spectrum to the chosen energy retention threshold at 12-hour, 120-hour, and 240-hour forecast lead times using ERA5 as reference. We evaluate thresholds of 0.3, 0.5, 0.7, and 0.9 to justify our threshold choice. A threshold of 0.9 disproportionately penalizes minor spectral deviations. Under this threshold, the effective resolution of NeuralGCM rapidly degrades to 2501.9 km, making it appear worse than purely data-driven MLWP models like Pangu-Weather, despite NeuralGCM's established capacity to resolve fine-scale dynamics. While lowering the threshold to 0.7 corrects this relative ranking between NeuralGCM and Pangu-Weather, it remains overly sensitive to localized spectral noise; for example, it assigns FuXi an effective resolution of over 6000 km. 

In contrast, a 0.3 threshold is too lenient. It makes it difficult to tell the models apart, as it fails to clearly show which models are smoothing out fine details as the lead time increases. Therefore, we select 0.5 as the default threshold for PhysMetrics.Weather. This value provides a balanced diagnostic: it is strict enough to clearly capture the decay of fine-scale variance in deterministic MLWP models, yet resilient enough to prevent minor spectral fluctuations from disproportionately skewing the metric.

\begin{figure}[h]
    \centering
    \includegraphics[width=0.95\linewidth]{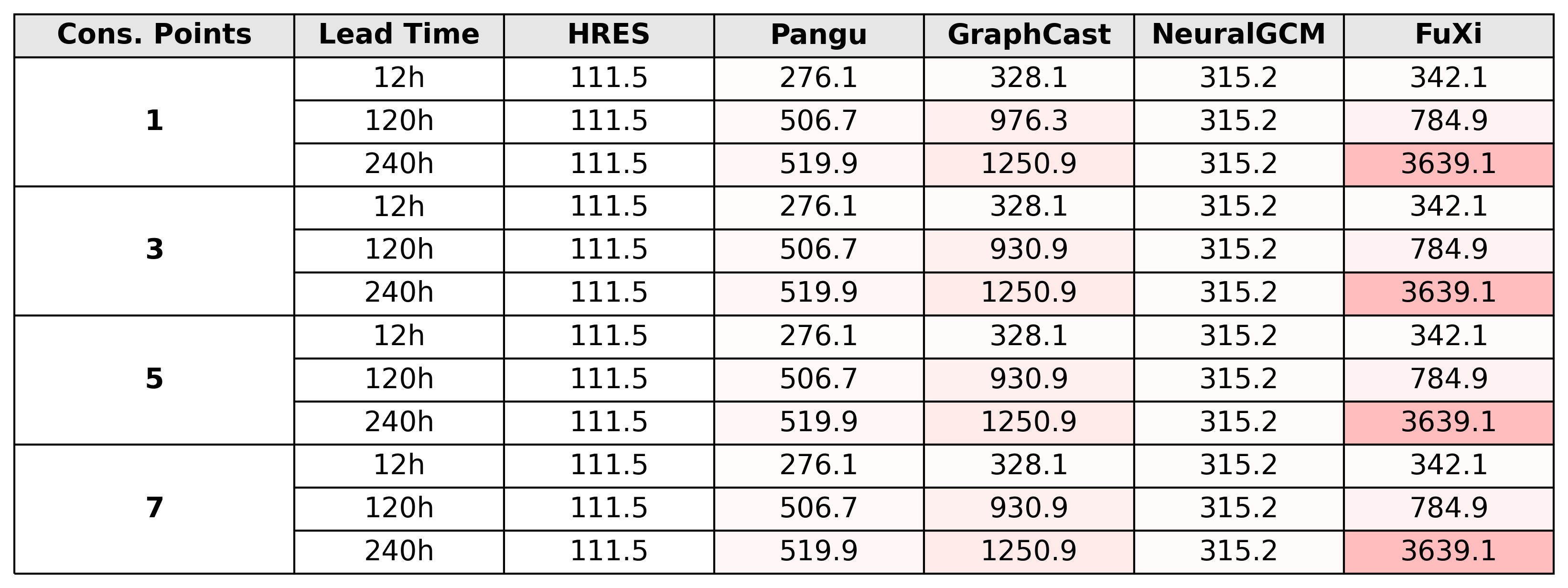}
    \caption{\textbf{Sensitivity of effective resolution to the consecutive wavenumber requirement with ERA5 reference.} Evaluated using a fixed retention threshold of 0.5. Native resolutions: 0.25° (111.5 km), except NeuralGCM (0.7°, 315.2 km). The metric demonstrates robustness.}
    \label{fig:cons}
\end{figure}

Figure \ref{fig:cons} evaluates the metric's sensitivity to the consecutive wavenumber requirement ($n$), keeping the threshold fixed at 0.5 and using ERA5 as reference dataset. As shown, the effective resolution is robust to the choice of $n$. The evaluated models yield nearly identical resolutions regardless of the consecutive point requirement, with only a minor deviation observed for GraphCast at the 120-hour lead time (976.3 km for $n=1$ versus 930.9 km for $n>1$). This confirms that the effective resolution is generally insensitive to this hyperparameter. Therefore, we keep $n=5$ as a default.

\subsection{850 hPa KE and q spectral analysis}\label{app:spectral_ablation}
\subsubsection{850 hPa KE spectrum}
\begin{figure*}[h]
    \centering
    
    \begin{subfigure}{\textwidth}
        \centering
        \includegraphics[width=0.95\linewidth]{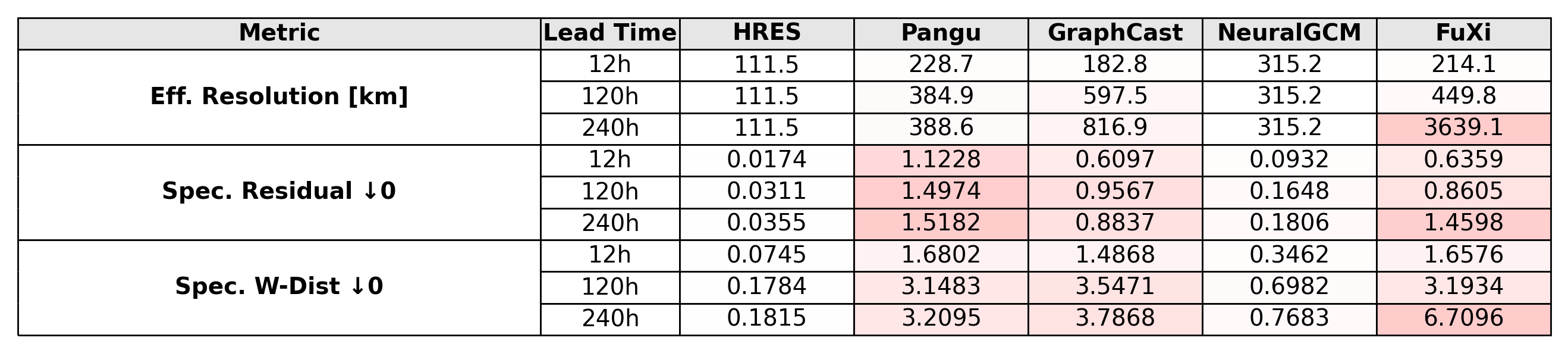} 
    \end{subfigure}
    
    \begin{subfigure}{0.95\textwidth}
        \centering
        \includegraphics[width=\linewidth]{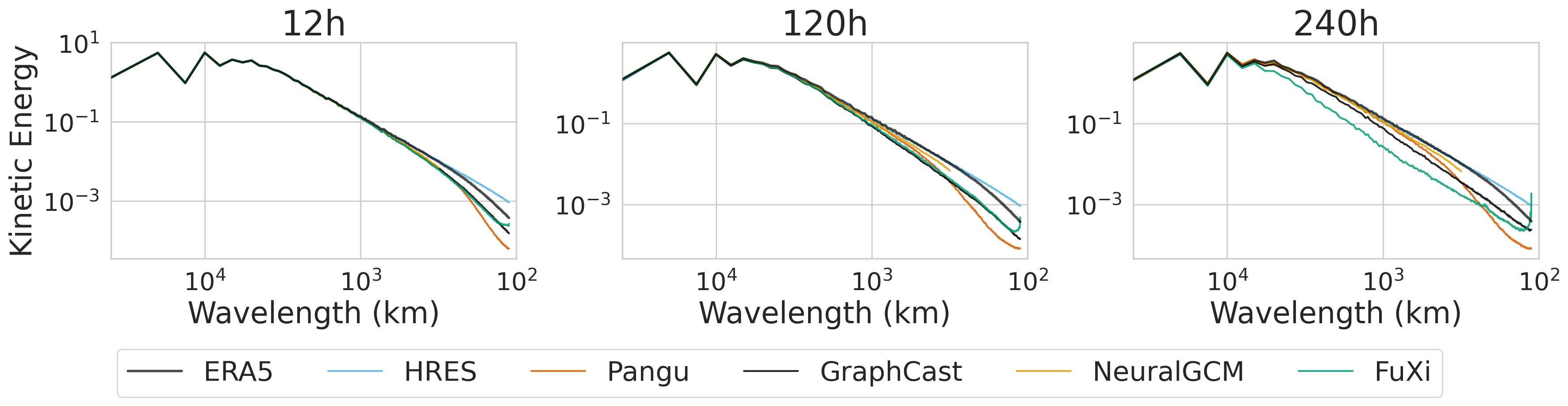}
    \end{subfigure}\hfill

    \caption{\textbf{KE spectra at 850 hPa with ERA5 reference.} \textbf{Top} Evaluative summary of spectral metrics. \textbf{Bottom} KE spectra at 12-h, 120-h, and 240-h lead times, respectively. Native resolutions: 0.25° (111.5 km), except NeuralGCM (0.7°, 315.2 km). While all models match the ERA5 reference at short lead times, extended forecasting periods reveal spatial smoothing (loss of fine-scale energy) in deterministic MLWP models. In contrast, differentiable hybrid architectures preserve the energy cascade.}
    \label{fig:combined_spectra_850}
\end{figure*}

Figure \ref{fig:combined_spectra_850} presents the KE spectral analysis evaluated at 850 hPa, complementing the 500 hPa analysis presented in the main text. At this lower altitude, the KE distribution features a lower secondary peak near the 10,000 km scale and a less smooth distribution of energy across spatial scales compared to the mid-troposphere, reflecting the increased influence of surface friction. The models reproduce these fine-scale variations relatively well.

The \textit{effective resolution} improves for Pangu-Weather and GraphCast at 850 hPa compared to 500 hPa. FuXi initially shows higher resolution at short lead times but degrades below its 500 hPa performance by the 240-hour mark. This apparent initial increase in resolution may be partially attributed to the metric's design, which requires the retention threshold to be violated for five consecutive wavenumbers. Furthermore, while all models achieve a lower (improved) \textit{spectral residual} at 850 hPa, their \textit{spectral divergence} generally gets worse, with the exception of the HRES analysis. This indicates that while the absolute errors decrease across the spectrum, the shape of the energy distribution diverges more significantly from the ERA5 reference.

\subsubsection{q spectrum analysis}
\begin{figure*}[h]
    \centering
    
    \begin{subfigure}{\textwidth}
        \centering
        \includegraphics[width=0.95\linewidth]{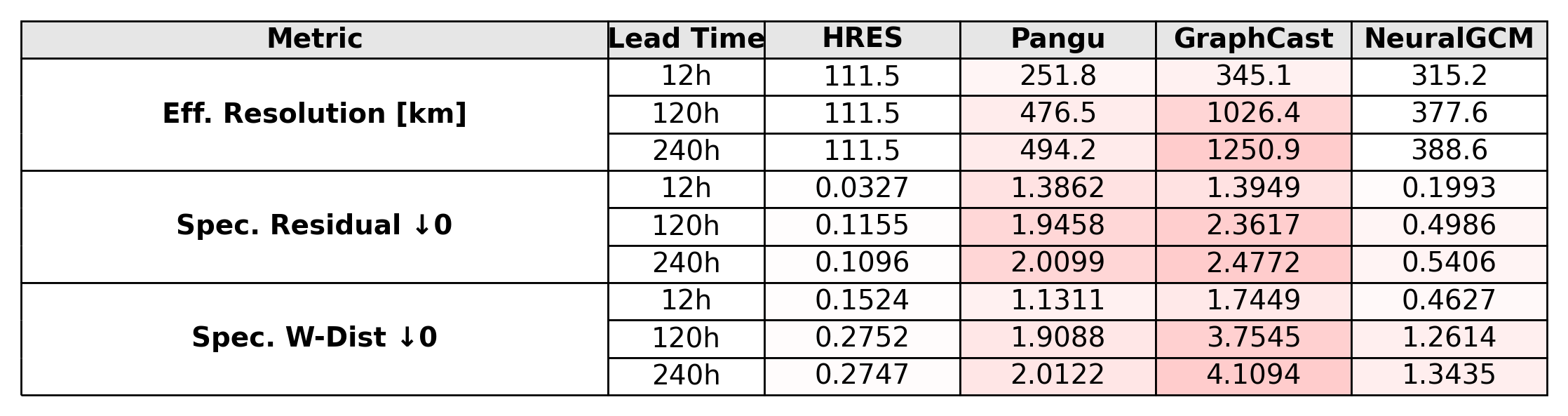} 
    \end{subfigure}

    \begin{subfigure}{0.95\textwidth}
        \centering
        \includegraphics[width=\linewidth]{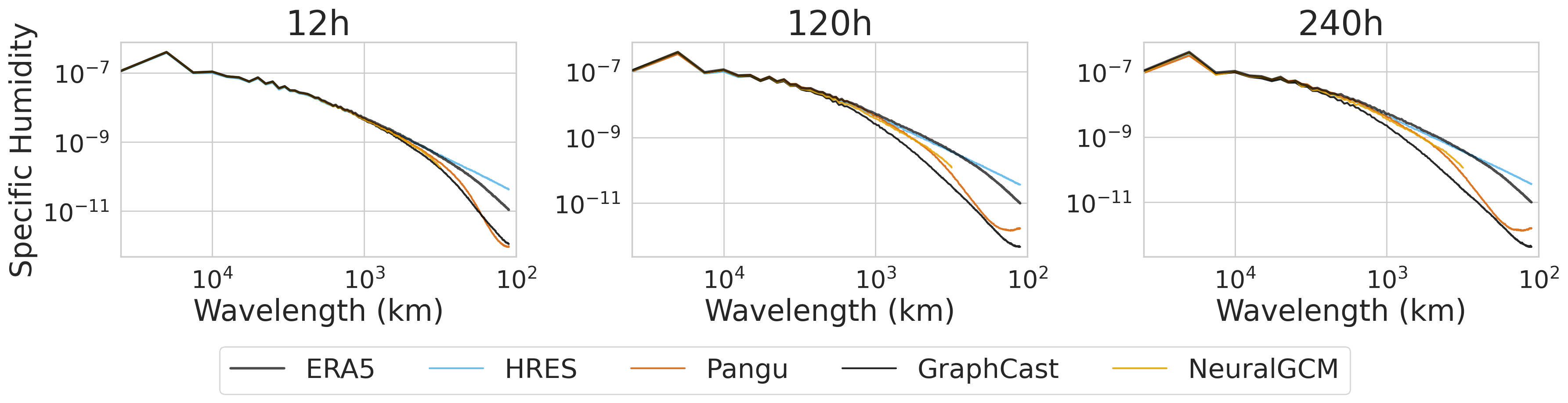}
    \end{subfigure}\hfill

    \caption{\textbf{KE spectra at the 500 hPa Q spectrum with ERA5 reference.} \textbf{Top} Evaluative summary of spectral metrics. \textbf{Bottom} Q spectra at 12-h, 120-h, and 240-h lead times, respectively. Native resolutions: 0.25° (111.5 km), except NeuralGCM (0.7°, 315.2 km). While all models match the ERA5 reference at short lead times, extended forecasting periods reveal spatial smoothing (loss of fine-scale energy) in deterministic MLWP models. In contrast, differentiable hybrid architectures preserve the energy cascade.}
    \label{fig:combined_spectra_q}
\end{figure*}

Figure \ref{fig:combined_spectra_q} shows the specific humidity ($q$) spectrum. The spectral power of the humidity distribution is several orders of magnitude lower (ranging from $10^{-12}$ to $10^{-7}$) compared to KE (ranging from $10^{-4}$ to $10^{1}$). Additionally, the reference humidity spectrum exhibits an uneven distribution of variance across spatial scales, lacking the smoother power-law shape seen in the KE spectrum.

When comparing these results to the 500 hPa KE analysis, the summary metrics reveal distinct changes in model behavior. Unlike its stable results across previous physical metrics, NeuralGCM exhibits a lower \textit{effective resolution} for humidity than it did for KE. Similarly, GraphCast retains less fine-scale variance for humidity compared to its KE performance, whereas Pangu-Weather's resolution actually improves. Regarding the \textit{spectral residual}, both HRES and NeuralGCM achieve lower absolute errors across the humidity spectrum than they did for KE. However, the \textit{spectral divergence} scores worsen for all data-driven models compared to their KE evaluations, indicating that their predicted humidity distributions deviate further from the ERA5 reference shape. Only the HRES analysis demonstrates a closer match to the reference distribution for humidity. This highlights the unique challenge MLWP models face in reproducing the highly variable, complex nature of atmospheric moisture.

\subsection{RMSE correlation}\label{app:RMSE}
\begin{figure}
    \centering
    \includegraphics[width=0.95\linewidth]{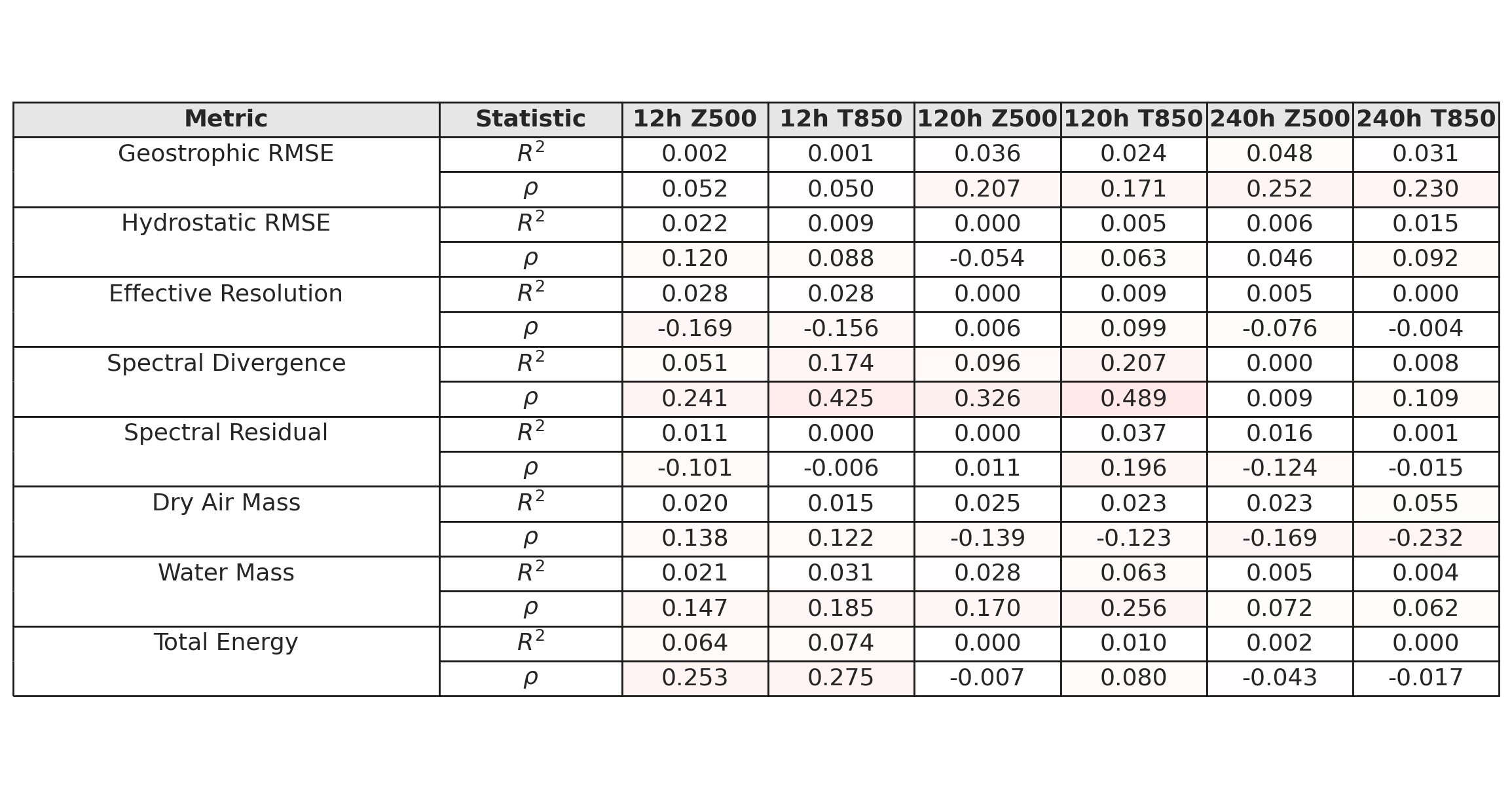}
    \caption{Coefficient of determination ($R^2$) and Spearman's rank correlation ($\rho$) between standard predictive error (RMSE) and physical consistency metrics. The correlations are computed using the daily predictions of Pangu-Weather over the year 2020. Results are shown for 12-h, 120-h, and 240-h forecast lead times, comparing the physical metrics against the standard RMSE for Geopotential at 500 hPa (Z500) and Temperature at 850 hPa (T850).}
    \label{fig:rmse_correlation}
\end{figure}

To demonstrate that PhysMetrics.Weather captures a distinct dimension of model performance, we evaluate the relationship between standard predictive skill and our physical metrics. Figure \ref{fig:rmse_correlation} presents the coefficient of determination ($R^2$) and Spearman's rank correlation ($\rho$) between the standard gridbox-weighted RMSE and the PhysMetrics.Weather metrics across the daily predictions for the year 2020. The consistently low $R^2$ and $\rho$ values reveal a weak correlation between a model's point-wise error and its physical fidelity. This confirms that PhysMetrics.Weather introduces independent diagnostics rather than acting as a proxy for standard RMSE.

\subsection{Ablation: Different surface pressure derivations}\label{app:sp}

\begin{figure}[htbp]
    \centering
    \includegraphics[width=0.95\linewidth]{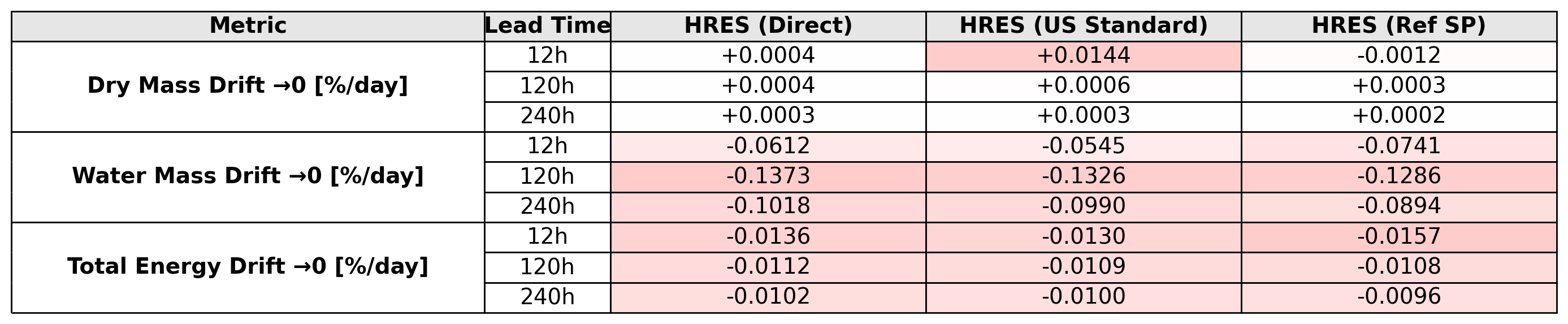}
    
    \includegraphics[width=0.95\linewidth]{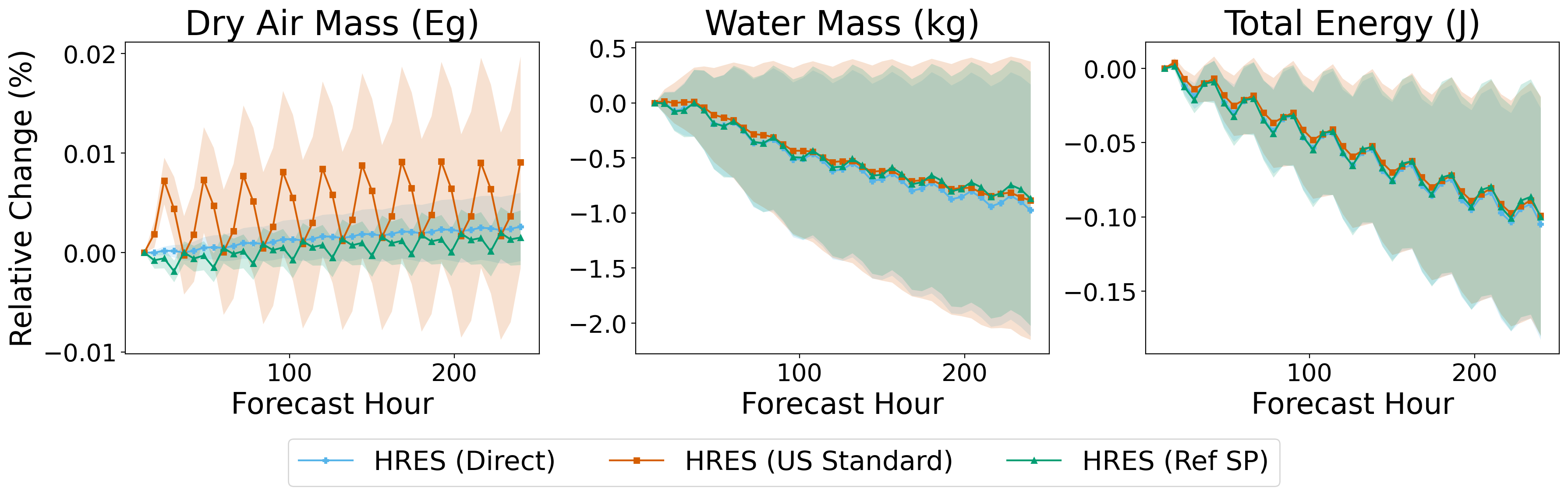}
    
    \caption{\textbf{Ablation study of global mass and energy conservation metrics using different surface pressure derivation strategies over a 240-hour forecast period.} An evaluation on IFS HRES using the IFS HRES analysis as reference dataset. \textbf{Top:} Summary of conservation metrics comparing the derivations. \textbf{Bottom:} Time series showing the relative and anomaly drifts. Using the derivation strategies yields long-term results highly similar to using the native surface pressure. Shading in time series indicates the $\pm 1$ standard deviation over 2020.}
    \label{fig:sp_abl}
\end{figure}

Figure \ref{fig:sp_abl} presents an ablation study evaluating the different surface pressure ($p_{s}$) derivation strategies applied to the IFS HRES model. For computational efficiency, the IFS HRES analysis is used as the reference baseline. 

We observe that relying on the U.S. Standard Atmosphere temperature profile introduces a visible diurnal cycle into the dry air mass time series. While this fluctuation noticeably impacts the \textit{relative drift} at the 12-hour lead time, the metric averages out these variations over extended time periods, ultimately yielding a drift comparable to that of the native $p_{s}$. In contrast, substituting with the reference $p_{s}$ induces a slight diurnal cycle in the opposite direction (reducing the relative change); however, this approach also remains similar to the directly available $p_s$. 

Importantly, the \textit{anomaly drift} for both global water mass and total energy is largely insensitive to the choice of surface pressure derivation. This stability is expected: for these quantities, $p_{s}$ primarily serves as the lower boundary condition for the vertical column integration, masking out pressure levels below the Earth's surface. In contrast, the global dry air mass calculation relies directly on $p_{s}$ to derive total atmospheric mass, making it more sensitive to short-term boundary fluctuations. 

Overall, the long-term conservation metrics remain  robust across all evaluated derivation methods. Consequently, we deem the standard atmosphere derivation strategy a safe and valid substitute when native surface pressure is unavailable.

\subsection{Ablation: Virtual temperature vs. dry temperature}\label{app:Tv}
In our balance analysis, we approximate virtual temperature with dry temperature ($T$) for FuXi, rather than using the standard formulation $T_v=T(1+0.6078q)$. To verify that this approximation does not significantly change the evaluation, we conduct an ablation study on all other evaluated models using the IFS HRES analysis as a reference dataset. When comparing the standard virtual temperature against the ablated dry temperature formulation, we observe that the results diverge by approximately $\pm 3$ m$^2$/s$^2$ across all models. Time-series analysis reveals that this difference is primarily driven by the virtual temperature being more sensitive to diurnal cycles than the dry temperature. Notably, the values in the time-series plot are nearly identical every 24 hours. Given these cyclical alignments and the relatively small magnitude of the overall divergence (up to $\pm 3$), we consider the use of dry temperature for FuXi to be a justified approximation within our framework

\begin{figure}[h]
    \centering
    \begin{minipage}{0.48\textwidth}
        \centering
        \includegraphics[width=\linewidth]{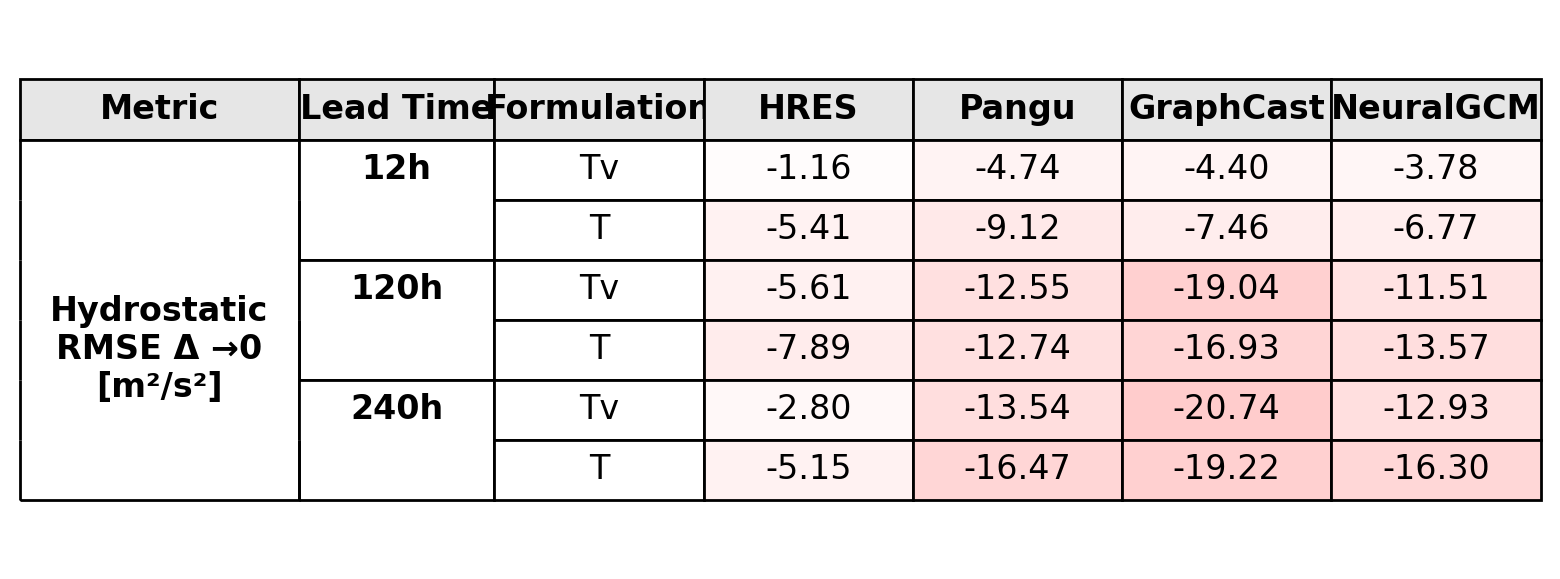}
    \end{minipage}\hfill
    \begin{minipage}{0.5\textwidth}
        \centering
        \includegraphics[width=\linewidth]{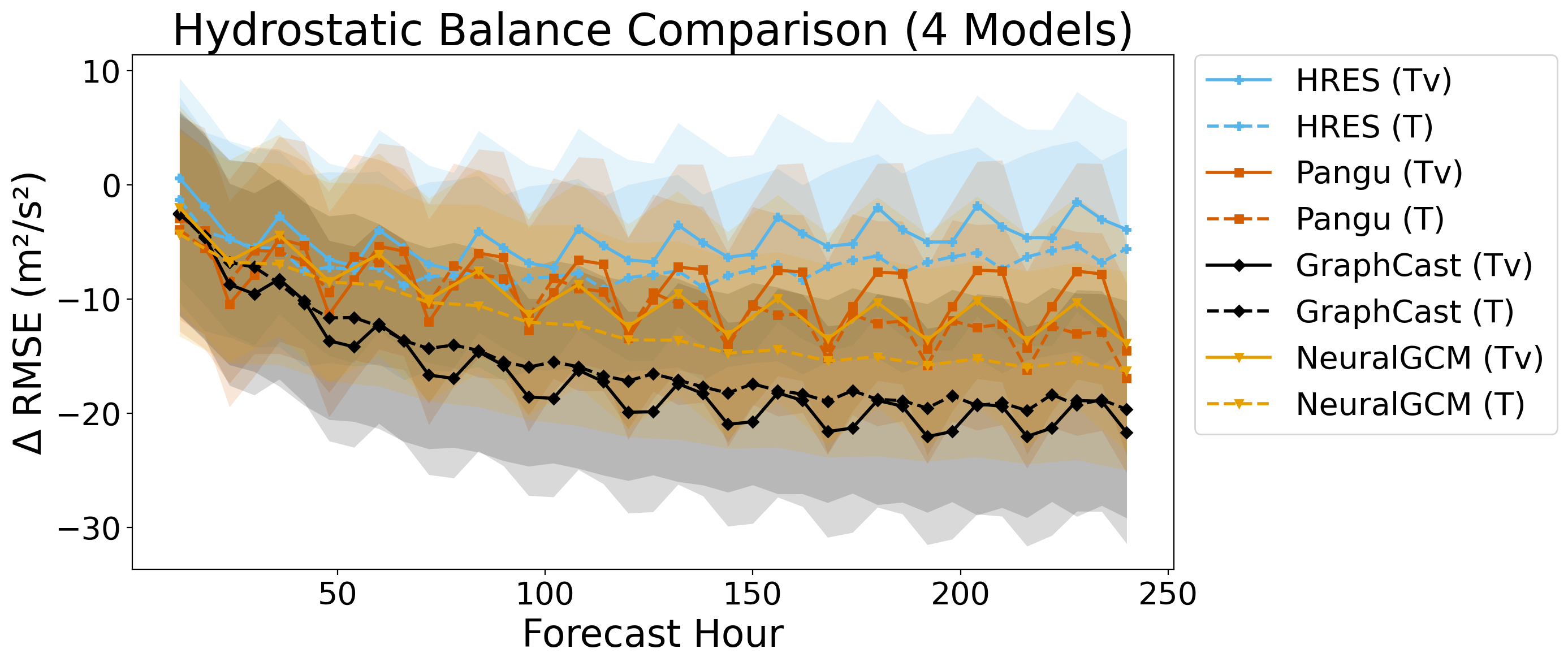}
    \end{minipage}
    
    \caption{\textbf{Ablation study of hydrostatic balance using dry versus virtual temperature over a 240-hour forecast period.} An evaluation of IFS HRES, Pangu, GraphCast, and NeuralGCM using the IFS HRES analysis as the reference dataset. \textbf{Left:} Summary of the hydrostatic balance metrics comparing virtual ($T_v$) and dry ($T$) temperature. \textbf{Right:} Time series showing the hydrostatic balance residuals. Using dry temperature yields results highly similar to virtual temperature, with minor divergences primarily driven by diurnal cycles. Shading in time series indicates the $\pm 1$ standard deviation over 2020.}
    \label{fig:hydrostatic_ablation}
\end{figure}

\section{Appendix D: Broader Context and Resources}
\subsection{Computational Costs}\label{app:compute}
All evaluations were performed on an anonymous high-performance computing cluster. For each experimental run, we used a single CPU node equipped with an AMD EPYC 7F32 (Rome) 3.7 GHz processor, specifically allocating 16 cores and 64 GB of RAM. Exact execution times for processing the 2020 dataset are detailed in Table \ref{tab:compute_resources}.

\begin{table}[H]
\centering
\caption{Execution times required for calculating each model's physical consistency metrics over the 2020 dataset. Because evaluations were performed against two different reference datasets (ERA5 in the main text and IFS HRES analysis in Appendix \ref{app:ifs}), processing times are reported separately for each comparison pipeline.}
\label{tab:compute_resources}
\begin{tabular}{@{}lcc@{}}
\toprule
\multirow{2}{*}{\textbf{Model}} & \multicolumn{2}{c}{\textbf{Execution Time}} \\ \cmidrule(l){2-3} 
 & \textbf{vs. ERA5} & \textbf{vs. IFS HRES} \\ \midrule
Pangu-Weather  & \textit{515m 16s} & \textit{480m 18s} \\
GraphCast      & \textit{906m 17s} & \textit{1166m 21s} \\
FuXi           & \textit{1301m 33s} & \textit{1177m 18s} \\
NeuralGCM      & \textit{303m 29s} & \textit{304m 25s} \\
IFS HRES       & \textit{511m 38s} & \textit{466m 10s} \\ \bottomrule
\end{tabular}
\end{table}

\subsection{Societal Impacts} \label{app:broader_impacts}

PhysMetrics.Weather provides a framework to evaluate the physical consistency of efficient MLWP models. Positively, this can help build operational trust, guide task-specific model selection, and help with safe MLWP adoption for downstream applications, such as extreme weather early warnings and climate risk assessments.

However, interpreting these results can come with risks. Because PhysMetrics.Weather focuses on global dynamics (constrained by current data availability), it does not assess all atmospheric variables or microphysical laws. High performance on our framework does not guarantee absolute physical consistency or no hallucinations. Relying solely on these metrics risks creating a false sense of security regarding a model's operational readiness. To mitigate this, PhysMetrics.Weather must be used alongside traditional meteorological validations and expert human oversight.

\subsection{Data and Code Availability} \label{app:licence}
\textbf{Code:} The code for the PhysMetrics.Weather framework, including all preprocessing, integration, and evaluation scripts, is available at \href{https://github.com/Emmakast/PhysMetrics.Weather}{GitHub} under the MIT License. 

\textbf{Datasets and Models:} This work relies on publicly available datasets and model weights provided by the WeatherBench 2 archive \citep{rasp_weatherbench_2024}. The WeatherBench 2 data is distributed under the Apache License 2.0. The underlying ERA5 reanalysis data \citep{Hersbach2023ERA5} is provided by the Copernicus Climate Change Service (C3S) and is subject to the Copernicus license terms.
\newpage

\end{document}